\documentclass{article}

\usepackage[preprint]{neurips_2026}
\usepackage{microtype}
\usepackage{graphicx}
\usepackage{subfigure}
\usepackage{booktabs} 
\usepackage[ruled, vlined]{algorithm2e}
\usepackage{soul}
\usepackage{ulem}
\usepackage{wrapfig}
\usepackage{subcaption}

\usepackage{booktabs}
\usepackage{xcolor}
\usepackage{colortbl}

\definecolor{lightgreen}{RGB}{220,245,220}
\definecolor{lightred}{RGB}{255,220,220}

\usepackage{hyperref}
\usepackage[table]{xcolor}
\usepackage{colortbl}

\usepackage{listings}
\usepackage{xcolor}

\usepackage{amsmath}
\usepackage{amssymb}
\usepackage{mathtools}
\usepackage{amsthm}
\usepackage{multirow}
\usepackage{enumitem}
\usepackage[capitalize,noabbrev]{cleveref}

\theoremstyle{plain}

\theoremstyle{definition}

\theoremstyle{remark}

\usepackage{color}

\title{Reinforced Collaboration in Multi-Agent Flow Networks}

\author{
Zheng Wang\thanks{Equal contribution.} \quad
Yuang Liu\footnotemark[1] \quad
Yangkai Ding \\
Huawei Technologies Co., Ltd. \\
\texttt{\{wangzheng155, dingyangkai\}@huawei.com}
}

\begin{document}

\maketitle

\newcommand{\good}[1]{\colorbox{green!15}{#1}}
\newcommand{\bad}[1]{\colorbox{red!15}{#1}}

\newcommand{\TBD}{}

\begin{abstract}
Multi-agent systems provide a powerful way to extend large language models (LLMs) by decomposing a complex task into specialized subtasks handled by different agents. However, their performance is often hindered by error propagation, arising from suboptimal workflow design or inaccurate agent outputs, which can propagate through the agent collaboration process and degrade final results.
To address the challenges, we present \texttt{MANGO} (Multi-Agent Network Gradient Optimization), a data-driven framework that organizes and refines agent collaboration via a flow network constructed from past successful workflows. \texttt{MANGO} integrates reinforcement learning and textual gradients to jointly optimize workflow paths and agent behaviors, while a skipping mechanism prevents redundant updates to well-optimized agents for improving efficiency. Extensive experiments on seven benchmarks show that \texttt{MANGO} achieves up to 12.8\% performance improvement over state-of-the-art baselines, enhances efficiency by 47.4\%, and generalizes effectively to unseen domains. Our code and datasets are publicly available on the \href{https://github.com/openJiuwen-ai/agent-store/tree/main/community/mango}{project page}.
\end{abstract}
\section{Introduction}
\label{sec:intro}

Multi-agent collaboration~\cite{zhuge_gptswarm_2024,zhang_maas_2025} has emerged as a promising paradigm for enhancing the problem-solving capabilities of large language models (LLMs). By decomposing a complex task into subtasks and assigning them to specialized agents, such systems exhibit collective intelligence in solving complex tasks~\cite{piatti2024cooperate,liu2023dynamic,du2023improving}.
Although multi-agent systems have demonstrated advantages, their architectures are intrinsically prone to \emph{error propagation}, as errors arising from incorrect workflow generation or hallucinated agent outputs can spread across the collaboration chain and affect final results. 
Despite recent efforts, existing methods still face several key challenges, as outlined below.

\begin{figure*}[t]
    \centering
    \includegraphics[width=\linewidth, trim={8.5mm 18cm 8.5mm 1.62cm}, clip]{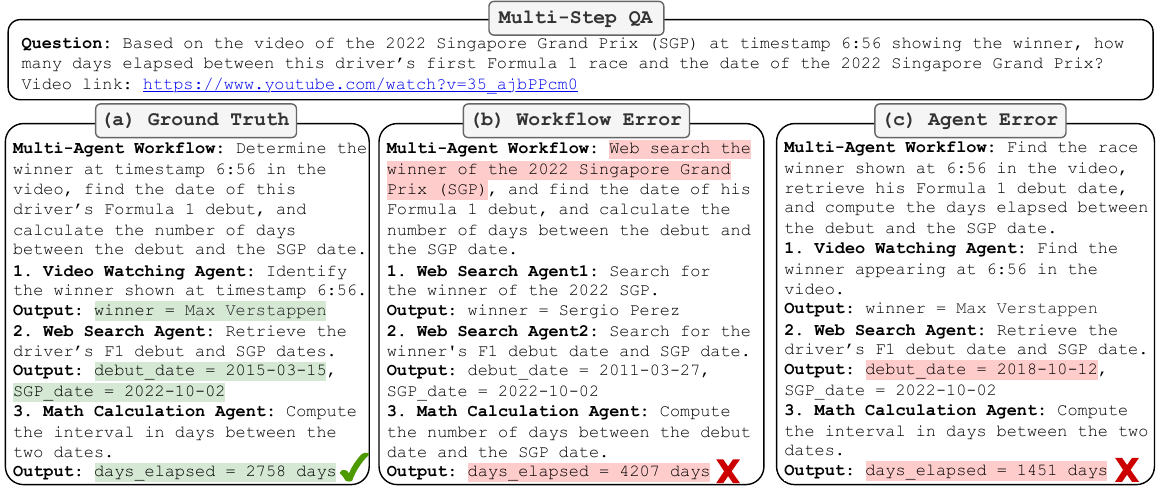}
    \caption{An example illustrating two types of errors in multi-agent collaboration, with hallucinated content highlighted: a \textbf{workflow error} occurs when a task is misplanned (e.g., identifying the winner via web search), and an \textbf{agent error} arises from incorrect execution (e.g., retrieving the wrong debut date). Both errors propagate through the multi-agent collaboration, degrading the final result.}
    \label{fig:introcase}
    \vspace{-5mm}
\end{figure*}

\textbf{Workflow Generation Error (C1).} This issue arises when an incorrect workflow is planned for a given task. As illustrated in Figure~\ref{fig:introcase}(a), solving a multi-step QA task requires (1) watching a video, (2) performing a web search, and (3) conducting a math calculation. In contrast, Figure~\ref{fig:introcase}(b) shows a failure case where the system mistakenly invokes a web search instead of watching the video. Such mistakes propagate through subsequent agent executions, ultimately affecting the final result.

Early multi-agent systems such as CAMEL~\cite{li2023camel}, AutoGen~\cite{wu2023autogen} and MetaGPT~\cite{hong2024metagpt} designed workflows based on various rules, including random dialogue~\cite{li2023camel}, cyclic dialogue~\cite{wu2023autogen}, manual or LLM specification~\cite{wu2023autogen}, and predefined scheduling~\cite{hong2024metagpt}. However, these methods often rely heavily on manual configurations, such as prompt engineering, which constrains the ability of multi-agent systems to quickly adapt to diverse domains and tasks.
Recent state-of-the-art agent systems such as ADAS~\cite{hu_adas_2024}, AgentSquare~\cite{shangagentsquare}, AFlow~\cite{zhang2025aflow}, and AgentSwift~\cite{li2025agentswift} explore automatically generating workflows using \emph{heuristic-driven strategies}. Specifically, ADAS~\cite{hu_adas_2024} adopts a greedy search in which a meta-agent iteratively creates, evaluates, and archives agents, continuing the search with the best updated archive. AFlow~\cite{zhang2025aflow} and AgentSwift~\cite{li2025agentswift} use an MCTS-based search that iteratively selects, expands, evaluates tree nodes and backpropagates the produced experiences. AgentSquare~\cite{shangagentsquare} adopts an evolutionary approach, generating new agents and exploring promising agent recombinations during the search process. However, the overall quality of these solutions depends heavily on the designed heuristics, making explicit optimization of performance uncertain.

\textbf{Single-agent Execution Error (C2).} This occurs when an agent generates inaccurate outputs (e.g., Figure~\ref{fig:introcase}(c) shows a web search agent retrieving a wrong debut date). Such local errors propagate through the workflow in multi-agent collaboration. For example, a mathematically correct agent still produces an incorrect result as it depends on the wrongly retrieved debut date.

Recent studies~\cite{yuksekgonul2025optimizing,pryzant2023automatic} explore textual gradient methods for mitigating LLM hallucination, which iteratively refine the system prompts using language feedback as differentiable gradients and improve outputs without updating model parameters. This idea has been extended to agent-based textual gradients, where a global gradient is estimated from the final result and backpropagated to optimize collaborating agents~\cite{zhang_maas_2025,hao2025chatllm,liu2023dynamic,hu2024self,zhou2024symbolic}. However, these approaches face two limitations: (1) the gradient is typically evaluated only at the final output, providing a weak supervision signal for intermediate agents in a long-chain collaboration, analogous to the gradient vanishing problem in neural network training; (2) backpropagating the gradient through all agents incurs high latency and redundant computation, e.g., if the video-watching agent in Figure~\ref{fig:introcase}(c) has already identified the correct winner, repeatedly re-optimizing its prompt provides no benefit and wastes resources. These issues highlight the need for more resource-efficient gradient update methods in multi-agent systems.

\textbf{New Solution.} In this paper, we study a novel flow network structure with a source and a sink node. The network is built from past successful workflows, with each node clustering similar operations from different workflows into a single agent. Within the flow network, workflow generation is formulated as finding a path from the source to the sink node, where the source node receives an input task and overall performance is evaluated at the sink. The performance is then converted into a textual gradient loss, which is backpropagated to optimize the prompts of agents along the selected path.
To address \textbf{C1}, our approach adopts a \ul{data-driven strategy} that leverages past experiences to learn dynamic workflow structures via process-supervised reinforcement learning. By using overall performance as the reward to guide workflow generation, it eliminates the need for manually designed heuristics in multi-agent collaboration. To address \textbf{C2}, we incorporate local gradient signals into the textual gradient for backpropagation. Additionally, we introduce a \ul{skipping node selection} strategy in the generated workflow, which allows the optimization process to bypass agents that are already well-optimized, thereby reducing the computational cost of repeated prompt updates. We refer to this solution as Multi-Agent Network Gradient Optimization (\texttt{MANGO}).


Although several existing studies~\cite{zhuge_gptswarm_2024, liu2023dynamic, zhang_maas_2025} model multi-agent systems using graph or network structures, our approach differs in two key aspects: (1) We introduce novel definitions of graph nodes and edges, specifically designed to mitigate error propagation by jointly optimizing workflow paths (edges) and agent outputs (nodes); (2) our method learns multi-agent collaboration patterns from past successful experiences and employs a skipping strategy to reduce computational overhead. 

In summary, we make the following contributions:

\begin{itemize}[noitemsep, topsep=0pt, leftmargin=6mm]
\item We identify a practical issue of error propagation in multi-agent collaboration and propose a novel flow network model that formulates this challenge as a joint optimization of edges and nodes in a data-driven manner. To our knowledge, this is the first work of its kind.
\item We propose \texttt{MANGO}, a framework that integrates reinforcement learning with textual gradients to jointly optimize overall performance. In addition, we design a skipping strategy to reduce computational costs.
\item Extensive experiments on seven benchmark datasets show that: (1) \texttt{MANGO} achieves up to 12.8\% improvement over SOTA baselines, (2) it reduces training and inference times by 41.5\% and 47.4\%, respectively, and (3) the trained \texttt{MANGO} model generalizes effectively to new domains and backbone LLMs not seen during training.
\end{itemize}

\section{Related Work}
\label{sec:related}

\textbf{LLM-based Autonomous Agents.} Recent work explores LLM-based autonomous agents~\cite{gao2025survey}, spanning single-agent and multi-agent systems. Representative single-agent frameworks such as AutoGPT~\cite{autogpt2023}, LangChain~\cite{langchain2022}, LlamaIndex~\cite{llamaindex2022}, and XAgent~\cite{xagent2023} enable LLMs to plan and execute user-specified tasks autonomously.

Multi-agent approaches extend this paradigm by enabling collaboration and role-based interactions among agents. Early works such as CAMEL~\cite{li2023camel}, Generalist Agents~\cite{park2023generalistagents}, ChatDev~\cite{qian2023chatdev}, AutoGen~\cite{wu2023autogen}, and MetaGPT~\cite{hong2024metagpt} rely on rule-based communication but suffer from hallucinations in task assignment and individual agent reasoning. More recent multi-agent frameworks, including ADAS~\cite{hu_adas_2024}, AFlow~\cite{zhang2025aflow},  AgentSquare~\cite{shangagentsquare}, MaAS~\cite{zhang_maas_2025}, GPTSwarm~\cite{zhuge_gptswarm_2024}, MAS-GPT~\cite{ye_masgpt_2025}, BiRouter~\cite{yang2026augmented}, and G-Designer~\cite{zhang_gdesigner_2024}, mitigate these limitations through enhanced planning and coordination. For example, AFlow~\cite{zhang2025aflow} leverages Monte Carlo Tree Search over code-defined workflows to automatically design adaptable multi-agent workflows, while MaAS~\cite{zhang_maas_2025} leverages an agentic supernet to dynamically generate task-specific architectures, allocating computational resources according to the difficulty and domain of the user query. However, the high performance of these methods relies on the designed heuristics (e.g., greedy search in ADAS, tree search in AFlow, evolutionary algorithm in AgentSquare, and agentic operator setups in MaAS). In contrast, we explore a data-driven paradigm that learns workflow generation from past experiences.

\textbf{Agentic Reinforcement Learning.} Agentic RL represents a paradigm shift from traditional RL approaches when applied to LLM-based agents~\cite{zhang2025landscape}. In this context, RL provides a foundation for evolving agent behaviors, transforming static, rule-based systems into adaptive, robust, and autonomous agents capable of diverse functions, including foundation model training~\cite{liao2025marft,liu2025llm,park2025maporl}, planning~\cite{song2024trial,wang2024voyager,lu2025pilotrl,wang2025instructrag, dang2025multi}, tool use~\cite{qian2025toolrl,feng2025retool}, memory management~\cite{tan2025prospect,yan2025memory,wang2024m}, reasoning~\cite{zeng2025simplerlzoo,yu2025dapo}, and self-improvement~\cite{she2025dupo,tian2025selfimprovement}. 
%
For example, Puppeteer~\cite{dang2025multi} proposes a puppeteer-style multi-agent framework where a RL–trained central orchestrator dynamically sequences and coordinates LLM agents to improve reasoning and reduce computational cost.
In this study, we explore a flow network framework for agentic systems that combines RL and textual gradients (TG), where RL selects nodes for TG optimization, and the optimized nodes, in turn, affect RL state construction. 


\textbf{Prompt Optimization.} Prompt optimization refines agent instructions to adapt its behavior without changing model weights. Early approaches relied on hand-crafted prompt designs, such as Chain-of-Thought~\cite{wei2022chain}, Tree-of-Thought~\cite{yao2023tree}, ReAct~\cite{yao2022react}, and Reflexion~\cite{shinn2023reflexion}. These methods improve prompt structures and enhance reasoning capabilities, but challenges remain in achieving more effective prompt optimization.
Recent prompt optimization techniques include EvoAgent~\cite{yuan2025evoagent}, which uses evolutionary algorithms to extend specialized agents into multi-agent systems, and SPO~\cite{xiang2025self}, a self-supervised framework for discovering effective prompts without external references.
%
TextGrad~\cite{yuksekgonul2025optimizing} formulates prompts as parameters in a differentiable program, refining them through natural-language gradient propagation. Building on this idea, DLPO~\cite{peng2025dlpo} extends textual gradient optimization with deep-learning-inspired techniques, including textual regularization and adaptive textual learning rates. 
However, these prompt optimization methods inherently incur high iterative computational costs. In this study, we investigate a skipping textual gradient strategy to selectively optimize agents, reducing computation for those already well-optimized.
\section{Preliminary}
\label{sec:preliminary}

\textbf{Workflow.} We denote a workflow $W^{(i)} = \langle e_1^{(i)}, e_2^{(i)}, \ldots \rangle$ as a sequence of operations executed by one or more LLMs to solve a task $T^{(i)}$ step by step, where $e_j^{(i)}$ denotes the operation produced at the $j$-th step of the $W^{(i)}$.

\textbf{Flow Network.} We define a flow network $\mathcal{G} = (\mathcal{V}, \mathcal{E}, s, t)$ as a directed graph with a source node $s$ and a sink node $t$, constructed from a set of workflows $\mathcal{W} = \{W^{(i)}\}_{i=1}^{|\mathcal{W}|}$. Let $\mathcal{V} = \{v_1, v_2, \ldots, v_{|\mathcal{V}|}\}$ denote a node set, where each node $v$ clusters similar operations $\{ e_j^{(i)} \mid 1 \le i \le |\mathcal{W}|, 1 \le j \le |W^{(i)}| \}$, and $\mathcal{E} \subseteq \mathcal{V} \times \mathcal{V}$ denotes the set of directed edges. Each node is modeled as an agent, with $v_i = (M_i, P_i, e_i^*, r_i^*)$ represented as a tuple comprising:
\begin{itemize}[noitemsep, topsep=0pt, leftmargin=6mm]
    \item \textbf{Model $M_i$}: the specific LLM executed at node $v_i$.
    \item \textbf{System Prompt $P_i$}: the system prompt for $M_i$ at node $v_i$, initially derived from the aggregated $\{e_j^{(i)}\}$ using $M_i$.
    \item \textbf{Demonstrated Experience $e_i^*$}: a representative experience of $\{e_j^{(i)}\}$, typically chosen as the cluster center, which captures the detailed behavior of this node.
    \item \textbf{Role Description $r_i^*$}: a summary derived from $\{e_j^{(i)}\}$ that describes the role of node $v_i$ in the multi-agent system.
\end{itemize}
The flow network construction is detailed in Section~\ref{sec:construction}.

\textbf{Problem Formulation.} Given a task $T$, we formulate the problem of solving this task as finding a path 
$\mathcal{P} = (v_1 = s, \ldots, v_{|\mathcal{P}|} = t)$, 
where each consecutive pair $(v_i, v_{i+1}) \in \mathcal{E}$. 
The task input is initiated at the source node $s$ and its performance is evaluated at the sink node $t$.

Each node $v_i$ is associated with a parametric prompt $\theta_i = (P_i, r_i^*)$, 
which defines how the node processes an incoming input. Formally, $f_i(x; \theta_i)$ denotes the output function (e.g., the response generated by an LLM $M_i$) at node $v_i$ for input $x$ under parameter $\theta_i$. We treat $P_i$ and $r_i^*$ as parametric components because they can be adjusted to control the behavior and output of the agent. Let
\begin{equation}
\Theta_{\mathcal{P}} = \{\theta_{v_1}, \theta_{v_2}, \ldots, \theta_{v_{|\mathcal{P}|}}\}    
\end{equation}
denote the set of prompt parameters along the path $\mathcal{P}$. The composition of the node functions along the path for a given task $T$ is then expressed as
\begin{equation}
\label{eq:pred}
    f_{\mathcal{P}}(T; \Theta_{\mathcal{P}}) 
    = f_{v_{|\mathcal{P}|}}(\cdot; \theta_{v_{|\mathcal{P}|}}) 
    \circ \cdots 
    \circ f_{v_2}(\cdot; \theta_{v_2}) 
    \circ f_{v_1}(T; \theta_{v_1}).
\end{equation}

Let $y$ be the target output of task $T$. Our objective is to minimize the discrepancy between the model prediction $f_{\mathcal{P}}(T; \Theta_{\mathcal{P}})$ and the ground truth $y$, as measured by a loss function $\mathcal{L}(\cdot, \cdot)$. Let $\Lambda(s, t)$ denote all valid paths from $s$ to $t$. The optimization problem can then be expressed as
\begin{equation}
    \min_{\mathcal{P} \in \Lambda(s, t),\ \Theta_{\mathcal{P}}} 
    \mathcal{L}\big(f_{\mathcal{P}}(T; \Theta_{\mathcal{P}}), y\big).
\end{equation}

The learning objective is to achieve a \textbf{joint optimization} of the optimal path $\mathcal{P}^*$ and its associated parameters $\Theta_{\mathcal{P}}^*$:

\begin{equation}
\label{eq:overall}
    (\mathcal{P}^*,\, \Theta_{\mathcal{P}}^*) 
    = \arg\min_{\mathcal{P} \in \Lambda(s,t),\, \Theta_{\mathcal{P}}}
    \mathcal{L}\big(f_{\mathcal{P}}(T; \Theta_{\mathcal{P}}),\, y\big).
\end{equation}

\section{Methodology}
\label{sec:method}

\begin{figure*}[htbp]
    \centering
    \includegraphics[width=\linewidth, trim={25.2mm 102.2cm 53.9cm 2.1cm}, clip]{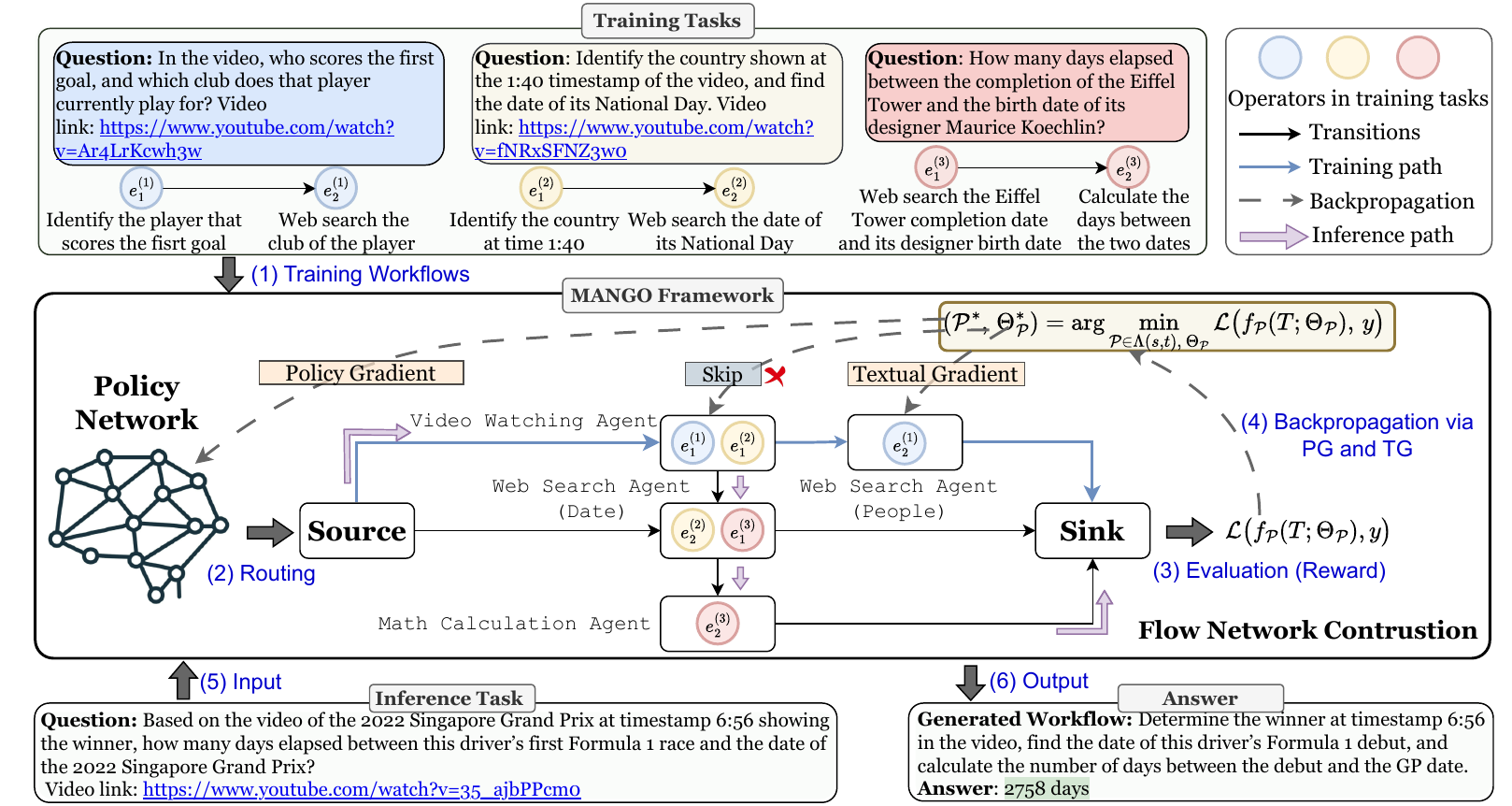}
    \vspace{-3mm}
    \caption{Overview of the \texttt{MANGO} framework. The flow network is constructed from past successful workflows, where a policy network routes each task from source to sink to find an optimal path. The framework jointly optimizes \textbf{path selection} via policy gradient (PG) and \textbf{prompt optimization} via textual gradient (TG), and \textbf{skips certain nodes} to reduce computational cost. During inference, the learned policy network generates an adaptive workflow to produce the final answer for a new task.}
    \label{fig:mango}
    \vspace{-6mm}
\end{figure*}

\subsection{MANGO Overview}
To solve the joint optimization problem of finding both the optimal path $\mathcal{P}^*$ and the corresponding node parameters $\Theta_{\mathcal{P}}^*$, we introduce a complementary update strategy.

For the node-level parameters $\Theta_{\mathcal{P}}$, which include the textual system prompt and role description, textual gradient descent~\cite{yuksekgonul2025optimizing} can be applied. The path $\mathcal{P}$, however, is discrete and thus optimized using a stochastic policy $\pi_\phi(\mathcal{P} \mid T)$, where the policy parameters $\phi$ are updated via the REINFORCE~\cite{williams1992simple}. 

Finally, the joint optimization can be expressed as a combined update of both components. The updates of node-level parameters $\Theta_{\mathcal{P}}$ and the path selection policy $\pi_\phi(\mathcal{P} \mid T)$ are mutually dependent: \textbf{the node parameters influence the path selection probabilities, while the sampled paths determine which node parameters are updated.} The overall framework is illustrated in Figure~\ref{fig:mango}.


\subsection{Flow Network Construction}
\label{sec:construction}

We construct the flow network in two steps, detailed in Algorithm~\ref{alg:flownetwork}.

\textbf{Step-1 (Generating Successful Workflows)}: For each training task, we generate candidate workflows using \texttt{GPT-4o-mini} (prompts in Figure~\ref{afig:prompts}) and verify them against ground-truth answers. We encourage brevity by selecting the shortest correct workflow, which is then used to construct the flow network. The robustness to low-quality workflows (e.g., from small LLMs) is analyzed in Appendix~\ref{asec:noise}.

\textbf{Step-2 (Inserting Workflows with a Threshold)}: We iteratively insert each operation $e^{(i)}_j$ in a workflow $W^{(i)}$ into the network $\mathcal{G}$. The process begins by inserting the first two operations ($e^{(i)}_1$ and $e^{(i)}_2$) to create two node sets ($v_1$ and $v_2$), ensuring adjacent operations are not placed in the same node to preserve their transitions. Each subsequent operation is inserted based on its similarity to existing node sets, defined as the maximum vector similarity between the operation and any operation within a node. If the similarity falls below a threshold $\delta$, a new node is created. Moreover, an operation is inserted into an existing node only if it is generated by the same LLM as that node, forming a multi-agent system in which each node represents an agent with a specific LLM. We note that this incremental insertion strategy groups semantically similar operations, preserves workflow structure, and enables the network to expand dynamically as additional workflows are incorporated.

We further leverage the LLM to derive the system prompt $P$, the cluster center $e^*$, and role description $r^*$ from the similar operation set $\{e_j^{(i)}\}$ to capture their shared behavior, collectively forming the context of the agent. We denote $\theta = (P, r^*)$ as the node parameter, which can be tuned to guide the agent toward optimal performance. The initial system prompt $P$ and the prompt for generating $r^*$ are provided in Figure~\ref{afig:prompts}.

\subsection{Edge Optimization with Reinforcement Learning}
\label{sec:edge}

\textbf{Overall Idea.} Given a constructed flow network $\mathcal{G}$ over historical workflows, our goal is to solve a new task $T$ by generating a workflow from the source node $s$ to the sink node $t$. The task is fed into the system at $s$, and its execution quality is evaluated at $t$. To achieve this, we adopt a plan-and-execute mode: a planner agent (\texttt{GPT-4o-mini}, and prompts in Figure~\ref{afig:planprompts}) at the source generates a high-level plan $\rho$ for solving the task. The plan consists of sequential steps $\rho_i$, and for each step we select an appropriate executor node from $\mathcal{G}$ according to the network structure. The planner is allowed to replan dynamically to improve future node selections. We formulate this decision-making process as a Markov Decision Process (MDP) and optimize it via reinforcement learning, as described below.

\textbf{Workflow Generation as an MDP.} To solve a task $T$, an agent incrementally constructs a workflow path $\mathcal{P}$ from the source $s$ to the sink $t$ within the flow network $\mathcal{G}$.

\ul{State.} At step $i$, the state $\mathbf{s}_i$ captures the relationship between the current plan step $\rho_i$ and the task $T$, along with the connected candidate node $v_j$ in $\mathcal{G}$ characterized by its role description $r^*_j$ and demonstrated experience $e^*_j$. That is
\begin{equation}
\label{eq:sims}
    \mathbf{s}_i = \{\mathrm{sim}(\rho_i, r^*_j), \mathrm{sim}(\rho_i, e^*_j), \mathrm{sim}(T, r^*_j), \mathrm{sim}(T, e^*_j)\},
\end{equation}
where $\mathrm{sim}(\cdot, \cdot)$ denotes a textual similarity measure, which can be implemented, for example, as the cosine similarity between text embeddings using \texttt{all-MiniLM-L6-v2}.

\ul{Action and Transition.} The action $a_i$ corresponds to selecting the next node $v_j$ to execute subtask $\rho_i$:

\begin{equation}
\label{eq:naiveaction}
a_i \in \mathcal{A}, \ \text{where} \ \mathcal{A} = \{ v_j \in \mathcal{V} \mid (v_i, v_j) \in \mathcal{E}\}.
\end{equation}

Executing $a_i$ induces a transition, i.e., $\mathbf{s}_{i+1} = \mathcal{T}(\mathbf{s}_i, a_i)$, where the next state $\mathbf{s}_{i+1}$ reflects the updated context after the node selection. We collect state-action pairs $(\mathbf{s}_i, a_i)$, evaluated by a policy network $\pi_\phi(a_i \mid \mathbf{s}_i)$ that selects the next node according to its probability distribution. This process is repeated sequentially until the workflow reaches the sink node $t$.

\ul{Reward.} The reward combines both process-level correctness and final task performance:
\begin{equation}
\label{eq:combinedreward}
r_i = \alpha\, r^{\text{(por)}}_i + (1-\alpha)\, r^{\text{(ror)}},
\end{equation}
where $r^{\text{(por)}}_i$ is a process-oriented reward indicating whether the selected node matches the ground-truth assignment in the training data ($r^{\text{(por)}}_i \in \{0,1\}$), and $r^{\text{(ror)}}$ is a result-oriented reward reflecting the final task outcome, e.g., accuracy for math tasks ($r^{\text{(ror)}} \in [0,1]$) evaluated at the sink node. The coefficient $\alpha \in [0,1]$ controls the trade-off between structural correctness and final performance.

\ul{Policy Optimization.}
The policy network $\pi_\phi(a \mid \mathbf{s})$ is optimized to maximize the expected cumulative reward using the REINFORCE algorithm~\cite{williams1992simple}:
\begin{equation}
\label{eq:edgeupdate}
\nabla_\phi J(\phi) = \mathbb{E}_{\pi_\phi} \Big[
\nabla_\phi \log \pi_\phi(a_i \mid \mathbf{s}_i) \, (R_i - b)
\Big],
\end{equation}
where $R_i = \sum_{t=i}^{|\mathcal{P}|} r_t$ is the accumulated reward from step $i$ to the end of the workflow, and $b$ is a baseline used to reduce variance. This objective encourages the policy to favor actions that generate effective workflows, optimizing both process correctness and end-to-end task performance.

We clarify the \ul{rationale for using RL} in two aspects: (1) RL enables exploration and recombination of flow substructures, allowing discovery of novel paths, whereas potential supervised methods mainly replicate observed paths and lack such flexibility; (2) RL optimizes long-horizon rewards, naturally modeling multi-step routing as an MDP. Additionally, our ablations in Table~\ref{tab:ablation} verify that RL outperforms heuristic routing methods, e.g., similarity ranking and top-candidate selection.

\subsection{Node Optimization with Textual Gradients}
\label{sec:node}

Let $\mathcal{P} = (v_1 = s, \ldots, v_{|\mathcal{P}|} = t)$ denote a sampled workflow path from the policy network. We then optimize the node-level textual parameters $\Theta_{\mathcal{P}} = \{(P_i, r_i^*)\}_{i=1}^{|\mathcal{P}|}$, where $P_i$ denotes the system prompt and $r_i^*$ denotes the role description associated with each node along the path. The objective is to leverage both the final task outcome (global signal) and intermediate execution feedback (local signal), ensuring that gradient signals do not vanish at earlier nodes when the workflow path is long.

\textbf{Global and Local Textual Gradient Update.}
At each node $v_i$, we apply textual gradient descent~\cite{yuksekgonul2025optimizing} using a combined gradient signal:
\begin{equation}
\nabla_{\Theta_i} \mathcal{L}
= \nabla_{\Theta}^{\text{(global)}} \mathcal{L}_\text{final} + \nabla_{\Theta_i}^{\text{(local)}} \mathcal{L}_i,
\end{equation}
where $\nabla_{\Theta}^{\text{(global)}} \mathcal{L}_\text{final}$ is computed on the sink node against the ground-truth result, and $\nabla_{\Theta_i}^{\text{(local)}} \mathcal{L}_i$ captures intermediate feedback (e.g., the output on $v_i$ conditioned on the preceding outputs). The parameters are then updated as:
\begin{equation}
\label{eq:nodeupdate}
\Theta_i \leftarrow
\Theta_i - \eta_\Theta \nabla_{\Theta_i} \mathcal{L},
\end{equation}
where $\eta_\Theta$ is the textual learning rate~\cite{peng2025dlpo}, which regulates the extent of prompt edits (e.g., number of modified sentences) at each step to ensure stable updates. We analyze the reliability and impact of local signals in Appendix~\ref{asec:signal}. The prompts for generating the signals are provided in Figure~\ref{afig:signal}.

\textbf{Mutual Dependency with Path Selection.} Updating $\Theta_i$ modifies the content of the state $\mathbf{s}_i$ (including the role $r_i$ and the generated plan step $\rho_i$), which directly affects the path selection policy $\pi_\phi(a_i \mid \mathbf{s}_i)$. Conversely, the sampled path determines which node parameters in $\Theta_{\mathcal{P}}$ are actually updated. This yields a mutually dependent optimization loop between parameter updates and path selection.

\begin{figure}[t]
\centering
\begin{minipage}[t]{0.55\linewidth}
\SetKwInOut{KwIn}{Require}
\begin{algorithm}[H]
\caption{Training Stage of \texttt{MANGO}}
\label{alg:training}
\KwIn{Flow network $\mathcal{G}$; training tasks $\mathcal{D}$}
  \For{epoch $=1,2,\ldots$}{
    \For{task $T \in \mathcal{D}$}{
      $\mathcal{P}\leftarrow (v_1=s)$, $i\leftarrow 1$, sample a plan $\rho$ at $s$
      
      \While{$v_i \ne t$}{
        Obtain $\rho_i$ and replan as needed at $s$


        $v_j \leftarrow a_i \sim \pi_\phi(a \mid \mathbf{s})$, $a_i \in \mathcal{A}$ with skipped \hfill $\triangleright$ \textcolor{blue}{Eq.}~\ref{eq:actionspace}
        
        $\mathcal{P} \leftarrow v_j$ (selected node from $\mathcal{G}$)
        
        $i\leftarrow j$
      }
      $y\leftarrow f_{\mathcal{P}}(T;\Theta_{\mathcal{P}})$ \hfill $\triangleright$ \textcolor{blue}{Eq.}~\ref{eq:pred}
      
      Calculate each reward $r_i$ with $y$ \hfill $\triangleright$ \textcolor{blue}{Eq.}~\ref{eq:combinedreward}
      
      Update $\pi_\phi$ along the path $\mathcal{P}$ \hfill $\triangleright$ \textcolor{blue}{Eq.}~\ref{eq:edgeupdate}
      
      Update $\Theta_{\mathcal{P}}$ via textual gradient \hfill $\triangleright$ \textcolor{blue}{Eq.}~\ref{eq:nodeupdate}\\
    }
  }
\textbf{Return} Trained $\pi_\phi$ and $\Theta$
\end{algorithm}
\end{minipage}
\hspace{3pt}
\begin{minipage}[t]{0.43\linewidth}
\SetKwInOut{KwIn}{Require}
\begin{algorithm}[H]
\caption{Inference Stage of \texttt{MANGO}}
\label{alg:inference}
\KwIn{Flow network $\mathcal{G}$;\\
trained $\pi_\phi$ and $\Theta$; \\
an input task $T$}

$\mathcal{P}\leftarrow (v_1=s)$

$i\leftarrow 1$

Sample a plan $\rho$ at $s$

\While{$v_i \ne t$}{

Obtain $\rho_i$ and replan as needed at $s$


$v_j \leftarrow a_i \sim \pi_\phi(a \mid \mathbf{s})$, $a_i \in \mathcal{A}$ with skipped \hfill $\triangleright$ \textcolor{blue}{Eq.}~\ref{eq:actionspace}

$\mathcal{P} \leftarrow v_{j}$ (selected node from $\mathcal{G}$)

$i\leftarrow j$\;
}
$y\leftarrow f_{\mathcal{P}}(T;\Theta_{\mathcal{P}})$ \hfill $\triangleright$ \textcolor{blue}{Eq.}~\ref{eq:pred}

\vspace{5pt}

\textbf{Return} Workflow $\mathcal{P}$ and the target output $y$ for task $T$
\end{algorithm}
\end{minipage}
\vspace{-6mm}
\end{figure}

\subsection{The \texttt{MANGO} Framework}
\label{sec:skip}

\textbf{Node Skipping for Computational Efficiency.} Optimizing workflow paths can be computationally expensive, primarily due to repeated LLM invocations to update the prompts at each node. Empirically, once a node's prompt is sufficiently optimized, further updates provide marginal gains. To address this, we introduce a \ul{skipping mechanism} that selectively skips certain nodes during optimization, thereby reducing computational overhead.
Instead of limiting action candidates to the immediate next nodes, the agent can choose from up to $K$-adjacent nodes; that is
\begin{equation}
\label{eq:actionspace}
a_i \in \mathcal{A}, \ \text{where} \ \mathcal{A} = \{ v_j \in \mathcal{V} \mid \text{dist}(v_i, v_j) \le K\}.
\end{equation}

Here, $\text{dist}(v_i, v_j)$ denotes the shortest-path distance from $v_i$ to $v_j$ (unit edge length). The parameter $K$ controls the agent's skip scope. Setting $K=1$ restricts the agent to its immediate neighbors. If one or more nodes are skipped, their corresponding outputs are filled using the steps from the training workflow. The rationale is to reuse ground-truth intermediate steps to supervise skipped nodes, improving training efficiency and stability without additional on-the-fly generation.


\textbf{Training and Inference Stages.} We optimize \texttt{MANGO} using workflows constructed over the network, enabling the agent to learn effective path routing through the policy network $\pi_\phi$ from correct workflows. We adopt a batch asynchronous training scheme, where tasks in each batch independently update the node parameters $\Theta_{\mathcal{P}}$, including system prompts and role descriptions.
During inference, given a new task, the learned policy network $\pi_\phi$ generates a workflow $\mathcal{P}$ for path selection. A skip operation is permitted, allowing a reached node to complete multiple sub-tasks corresponding to skipped nodes. All node parameters $\Theta_{\mathcal{P}}$ are pre-optimized during training and used directly at inference without further updates, thereby avoiding additional online computation. Algorithm~\ref{alg:training} and Algorithm~\ref{alg:inference} summarize the overall training and inference stages.
\section{Experiments}
\label{sec:exp}

\subsection{Experimental Setup}
\label{sec:setup}

\textbf{Datasets and Evaluation Metrics.} We conduct extensive experiments to demonstrate the capabilities of \texttt{MANGO} across seven widely-used benchmarks:
(1) HumanEval~\cite{chen2021evaluating} and MBPP~\cite{austin2021program} for \textbf{code generation};
(2) MATH~\cite{hendrycks2021measuring} and GSM8K~\cite{cobbe2021training} for \textbf{math reasoning};
(3) DROP~\cite{dua2019drop} for \textbf{reading comprehension};
(4) MMLU~\cite{hendrycksmeasuring} for \textbf{multi-task} problem solving; and
(5) GPQA~\cite{rein2024gpqa} for solving hard graduate-level \textbf{science questions}.

For evaluation metrics, following prior works~\cite{zhang2025aflow, hu_adas_2024, zhang_maas_2025, zhuge_gptswarm_2024}, we report \textbf{pass@1} for HumanEval and MBPP, \textbf{Accuracy (\%)} for MATH, GSM8K, MMLU, and GPQA, and \textbf{F1 Score} for DROP. In addition, we follow the dataset splits (train and test) used in~\cite{zhang2025aflow, zhang_maas_2025}, and detailed dataset statistics are provided in Appendix~\ref{asec:datasets}. We note that reported experimental results are \ul{statistically significant}, as verified by a t-test with $p<0.05$.

\textbf{Baselines.} We compare \texttt{MANGO} against two categories of agentic baselines:
(1) \textbf{single-agent methods}, including Chain-of-Thought~\cite{wei2022chain}, Self-Consistency CoT (5 samples)~\cite{wangself}, and Self-Refine (up to 3 refinement rounds)~\cite{madaan2023self};
(2) \textbf{hand-crafted or autonomous multi-agent methods}, including MultiPersona~\cite{wang2023unleashing}, LLM-Debate~\cite{du2023improving}, DyLAN~\cite{liu2023dynamic}, Plan-and-Execute~\cite{langchain_planning_agents_2024}, GPTSwarm~\cite{zhuge_gptswarm_2024}, ADAS~\cite{hu_adas_2024}, AgentSquare~\cite{shangagentsquare}, AFlow~\cite{zhang2025aflow}, and MaAS~\cite{zhang_maas_2025}.
Detailed configurations are in Appendix~\ref{asec:baseline}.

\textbf{Implementation Details.} Implementation details of \texttt{MANGO} and baselines are in Appendix~\ref{asec:implementation}.

\begin{table*}[t]
\centering
\caption{Effectiveness results with \texttt{GPT-4o-mini} as the base LLM. Best results are bold and shaded in green, and best baselines are underlined.}
\vspace{-2mm}
\label{tab:effectiveness_baselines}
\setlength{\tabcolsep}{3.5pt}
\begin{tabular}{l|ccccccc}
\hline
\textbf{Method}            & \textbf{HumanEval} & \textbf{MBPP} & \textbf{MATH} & \textbf{GSM8K} & \textbf{DROP} & \textbf{MMLU} & \textbf{GPQA} \\ \hline
IO (direct LLM invocation) &         87.08        &     71.83       &      46.29      &     87.45        &     68.25      &        63.44   &     28.79     \\ \hline
CoT~\cite{wei2022chain} &         88.13        &      71.83      &      46.40      &      87.10      &        78.50     &       65.43    &      29.29      \\
SC (CoT×5)~\cite{wangself} &        88.60       &       73.60     &      47.91      &       87.57      &       78.73     &      65.90     &      30.81      \\
Self-Refine~\cite{madaan2023self} &       87.79          &      69.79      &       46.09     &      89.63      &    70.27        &       63.50     &       31.82     \\ \hline
MultiPersona~\cite{wang2023unleashing} &         88.32        &     73.19       &       45.43     &      87.50       &      74.38      &     65.94      &       31.82    \\
LLM-Debate~\cite{du2023improving} &        88.68         &      70.29      &     48.54       &      89.47       &      60.63         &      65.64      &       31.40     \\
DyLAN~\cite{liu2023dynamic} &         90.42        &      77.30      &      48.63     &      89.98       &     80.29      &      80.32    &       36.87    \\
Plan-and-Execute~\cite{langchain_planning_agents_2024} &         88.71        &      70.30      &     48.54       &         89.52    &      80.13     &     80.33       &   37.37       \\
GPTSwarm~\cite{zhuge_gptswarm_2024} &        89.32         &      77.43      &      47.88     &       89.14      &      79.77    &      78.34      &        38.38        \\
ADAS~\cite{hu_adas_2024} &        84.19        &      68.13     &      43.18      &       86.12      &       76.63     &      69.60      &      34.57      \\
AgentSquare~\cite{shangagentsquare} &          89.08          &        78.46       &        48.51       &         87.62       &     78.38     &   78.01        &      37.88      \\
AFlow~\cite{zhang2025aflow} &         90.93      &       81.67     &     51.28      &      91.16      &       \underline{80.63}     &    \underline{80.52}     &      38.38     \\
MaAS~\cite{zhang_maas_2025} &         \underline{92.85}        &      \underline{82.17}      &      \underline{51.82}      &      \underline{92.30}       &    80.25    &    80.30    &    \underline{39.39}   \\ \hline
\rowcolor{green!5}
\texttt{MANGO} &         \textbf{95.42}         &       \textbf{85.63}      &       \textbf{58.44}      &       \textbf{94.12}      &       \textbf{84.75}     &       \textbf{81.97}     &       \textbf{40.40}      \\ 
\texttt{MANGO} (Skip-2) &94.66 &83.58 &54.32 &92.70 &82.91 &80.73 &38.89\\
\hline
\end{tabular}
\vspace{-2mm}
\end{table*}

\begin{table*}[t]
\centering
\caption{Efficiency comparison between \texttt{MANGO} and baseline methods on MATH.}
\vspace{-2mm}
\setlength{\tabcolsep}{2.2pt}
\small
\label{tab:efficiency}
\begin{tabular}{l|cccc|cccc|c}
\hline
\multirow{3}{*}{\textbf{Method}} & \multicolumn{4}{c|}{\textbf{Training}}                                                                                                                                                                                                                      & \multicolumn{4}{c|}{\textbf{Inference}}                                                                                                                                                                                                                     & \textbf{Overall}                                             \\ \cline{2-10} 
                        & \begin{tabular}[c]{@{}c@{}}Prompt \\ token\end{tabular} & \begin{tabular}[c]{@{}c@{}}Completion\\ token\end{tabular} & \begin{tabular}[c]{@{}c@{}}Total\\cost (\$)\end{tabular} & \begin{tabular}[c]{@{}c@{}}Time\\ (min)\end{tabular} & \begin{tabular}[c]{@{}c@{}}Prompt \\ token\end{tabular} & \begin{tabular}[c]{@{}c@{}}Completion\\ token\end{tabular} & \begin{tabular}[c]{@{}c@{}}Total\\cost (\$)\end{tabular} & \begin{tabular}[c]{@{}c@{}}Time\\ (min)\end{tabular} & \begin{tabular}[c]{@{}c@{}}Acc.\\ (\%)\end{tabular} \\ \hline
SC (CoT×5)             & - & - & - & - &524, 755&1, 784, 658&1.150&22&47.91\\ \hline
GPTSwarm&21, 325, 266& 6, 369, 884&7.021&129&3, 105, 571&788, 273&0.939&30&47.88\\
AFlow&33, 831, 239& 29, 051, 840&22.506&184&2, 505, 944& 2, 151, 931&1.667&23&51.28\\
MaAS&\cellcolor{green!5} \textbf{3, 052, 159}&2, 380, 505&1.886&53&\cellcolor{green!5} \textbf{1, 311, 669}&853, 116&0.709&19&51.82\\ \hline
\texttt{MANGO}&4, 221, 530&1, 094, 621&1.290&42&2, 858, 689&843, 436&0.935&12&\cellcolor{green!5}\textbf{58.44}\\
\texttt{MANGO} (Skip-2)&3, 558, 837&993, 553&1.130&33&2, 297, 606&756, 828&0.799&11&54.32\\
\texttt{MANGO} (Skip-3)&3, 240, 869&\cellcolor{green!5}\textbf{942, 372}&\cellcolor{green!5}\textbf{1.052}&\cellcolor{green!5}\textbf{31}&1, 944, 048&\cellcolor{green!5}\textbf{671, 269}&\cellcolor{green!5}\textbf{0.694}&\cellcolor{green!5}\textbf{10}&53.09\\ \hline
\end{tabular}
\vspace{-4mm}
\end{table*}

\subsection{Experimental Results}
\label{sec:result}

\textbf{(1) Effectiveness Comparison with Baseline Methods.} As shown in Table~\ref{tab:effectiveness_baselines}, we evaluate existing agent systems, including both single-agent and multi-agent methods in standard benchmark test sets following~\cite{zhang_maas_2025}.
Overall, \emph{\texttt{MANGO} consistently achieves the best performance across all task domains}, surpassing the best baselines, for example, improving {\TBD{over MaAS by 12.8\% on MATH and over AFlow by 5.1\% on DROP}}. These gains come from \texttt{MANGO} learning from past workflows and jointly optimizing workflow generation and single-agent execution, and remain superior even under the Skip-2 setting. In addition, \texttt{MANGO} can incorporate frontier models (e.g., GPT-5), further improving performance as base model capabilities increase (Appendix~\ref{asec:gpt5}).


\textbf{(2) Efficiency and Training/Inference Costs.} Following~\cite{zhang_maas_2025}, we conduct a cost analysis of \texttt{MANGO} against strong single-agent and multi-agent baselines. In Table~\ref{tab:efficiency}, \emph{\texttt{MANGO} with node skipping achieves the best cost-efficiency}, with lower token usage, API cost, and runtime while maintaining top accuracy.
For example, by enabling three-node skipping (Skip-3) during the flow network traversal, \texttt{MANGO} attains the lowest API cost (\$0.15 per million prompt token and \$0.6 per million completion token), while reducing training time by {\TBD{41.5\%}} and inference time by {\TBD{47.4\%}} compared to MaAS, and maintaining the highest accuracy. These results indicate that the proposed skipping mechanism effectively learns to skip already-optimized nodes, improving overall efficiency.

\begin{wraptable}{r}{0.6\linewidth}
\vspace{-4mm}
\centering
\caption{Effectiveness of \texttt{MANGO} across LLMs. The model is trained with \texttt{GPT-4o-mini} (\texttt{G}) and tested with different LLMs of \texttt{Qwen-2.5-72B} (\texttt{Q}) and \texttt{Llama-3.1-70B} (\texttt{L}).}
\vspace{-2mm}
\setlength{\tabcolsep}{2.5pt}
\label{tab:effectiveness_llms}
\small
\begin{tabular}{l|clcl|clcl}
\hline
\textbf{Dataset}               & \multicolumn{4}{c|}{\textbf{HumanEval}}                                                                             & \multicolumn{4}{c}{\textbf{MATH}}                                                                             \\ \hline
\multirow{2}{*}{LLMs} & \multicolumn{2}{c|}{Zero-shot}                              & \multicolumn{2}{c|}{Few-shot}                & \multicolumn{2}{c|}{Zero-shot}                              & \multicolumn{2}{c}{Few-shot}           \\ \cline{2-9} 
                      & \texttt{G}$\rightarrow$\texttt{Q}      & \multicolumn{1}{l|}{\texttt{G}$\rightarrow$\texttt{L}} & \texttt{G}$\rightarrow$\texttt{Q}      & \texttt{G}$\rightarrow$\texttt{L}       & \texttt{G}$\rightarrow$\texttt{Q}      & \multicolumn{1}{l|}{\texttt{G}$\rightarrow$\texttt{L}} & \texttt{G}$\rightarrow$\texttt{Q}      & \texttt{G}$\rightarrow$\texttt{L} \\ \hline
AFlow                & \multicolumn{1}{l}{88.55} & \multicolumn{1}{l|}{87.02}                & \multicolumn{1}{l}{90.08} & 87.79            & \multicolumn{1}{l}{63.17} & \multicolumn{1}{l|}{33.54}& \multicolumn{1}{l}{63.99} & 35.39    \\
MaAS                  &90.14& \multicolumn{1}{l|}{85.26}&90.84& \multicolumn{1}{c|}{87.02} &69.35& \multicolumn{1}{c|}{42.97}&69.55&45.06\\ \hline
\rowcolor{green!5}
\texttt{MANGO}                 & \multicolumn{1}{l}{91.60} & \multicolumn{1}{l|}{87.79}                & \multicolumn{1}{l}{92.37} &88.96& \multicolumn{1}{l}{69.75} & \multicolumn{1}{l|}{44.03}                & \multicolumn{1}{l}{69.96} &     48.15       \\ \hline
\end{tabular}
\vspace{-2mm}
\end{wraptable}
\textbf{(3) Effectiveness Comparison across LLMs.} We study the robustness of multi-agent systems across different LLM backbones. In Table~\ref{tab:effectiveness_llms}, \emph{\texttt{MANGO} demonstrates strong LLM-agnostic robustness}. Both \texttt{MANGO} and baselines are trained with \texttt{GPT-4o-mini} and transferred to \texttt{Qwen-2.5-72B} and \texttt{Llama-3.1-70B} under zero-shot and few-shot settings (fine-tuned on 20\% of the training data).
\texttt{MANGO} consistently outperforms the best baselines, with gains of up to {\TBD{2.5\%}} (zero-shot) and {\TBD{6.9\%}} (few-shot), due to its high-quality, model-agnostic workflows that better guide different LLMs.

\begin{wraptable}{r}{0.6\linewidth}
\vspace{-4mm}
\caption{In-domain and cross-domain transferability of \texttt{MANGO}, where MA$\rightarrow$GS (or HE) indicates that the model is trained on MATH and evaluated on GSM8K (or HumanEval).}
\vspace{-2mm}
\setlength{\tabcolsep}{4pt}
\label{tab:transfer}
\begin{tabular}{l|cccc}
\hline
\textbf{Transfer} & \textbf{MA$\rightarrow$GS} & \textbf{GS$\rightarrow$MA} & \textbf{MA$\rightarrow$HE} & \textbf{HE$\rightarrow$MA}\\\hline
AFlow         &91.95&49.39&89.13&47.15\\
MaAS          &92.80&51.02&91.15&50.27\\ \hline
\rowcolor{green!5}
\texttt{MANGO}&93.08&54.53&92.37&54.12\\ \hline
\end{tabular}
\vspace{-3mm}
\end{wraptable}
\textbf{(4) In-Domain and Cross-Domain Transferability Study.} We evaluate model transferability by directly applying a model trained on one task to another task, either within the same domain or across different domains. As shown in Table~\ref{tab:transfer}, \emph{\texttt{MANGO} shows strong in-domain and cross-domain generalization}, outperforming the best baselines {\TBD{(MaAS) by up to 7.7\%}}. These gains stem from the designed state features (similarity measures) which capture transferable patterns across domains.

\begin{wraptable}{r}{0.55\linewidth}
\vspace{-4mm}
\centering
\caption{Ablation study on the impact of reinforcement learning (RL) and textual gradients (TG) on MATH.}
\vspace{-2mm}
\label{tab:ablation}
\setlength{\tabcolsep}{3pt}
\small
\begin{tabular}{l|ccc}
\hline
\textbf{Metric}                   & \textbf{Acc.} & \textbf{Train} & \textbf{Test} \\ \hline
\texttt{MANGO}           &58.44&42&12\\ \hline
w/o RL by $\mathrm{sim}(\rho_i, r^*_j)+\mathrm{sim}(T, r^*_j)$  &54.94&23&10\\
w/o RL by $\mathrm{sim}(\rho_i, e^*_j)+\mathrm{sim}(T, e^*_j)$  &52.67&23&10\\
w/o RL by LLM invocation  &51.23&44&16\\ \hline
RL w/o $\mathrm{sim}(\rho_i, r^*_j)$ &54.94 &41 &11\\
RL w/o $\mathrm{sim}(\rho_i, e^*_j)$ &56.79 &41 &11\\
RL w/o $\mathrm{sim}(T, r^*_j)$ &55.97 &41 &11\\
RL w/o $\mathrm{sim}(T, e^*_j)$ &56.17 &41 &11\\ \hline
w/o TG             &48.97&11&8\\
TG w/o local signals             &54.73 &32 &11\\
\hline
\end{tabular}
\vspace{-3mm}
\end{wraptable}
\textbf{(5) Ablation Study.} We conduct ablations on two key components of \texttt{MANGO}: (1) path selection without RL (using similarity-based routing or LLM calls), and (2) parameter optimization without textual gradients. As shown in Table~\ref{tab:ablation}, prompt optimization contributes the largest gain of 19.3\%, with local signals contributing 6.8\%, highlighting the impact of error propagation in multi-agent systems.
Similarity-based routing improves efficiency, while LLM-based selection increases latency. We also remove each state similarity feature and observe consistent performance drops from 58.44, showing that all features contribute to routing decisions.





\section{Conclusion}
\label{sec:conclusion}


We study error propagation in multi-agent collaboration, identifying workflow generation and single-agent execution errors. We propose \texttt{MANGO}, a data-driven framework that learns a flow network and integrates RL, textual gradients, and a skipping mechanism for efficient optimization. Experiments show that \texttt{MANGO} achieves superior performance while reducing computational cost.

\normalem
\bibliographystyle{plain}
\bibliography{ref}


\newpage
\appendix



\section{Experimental Details}

\subsection{Dataset Statistics}
\label{asec:datasets}

We follow the train–test splits used in~\cite{zhang2025aflow, zhang_maas_2025} and report the dataset statistics in Table~\ref{atab:statistics}.
\begin{table}[h]
\centering
\caption{Dataset Statistics.}
\label{atab:statistics}
\begin{tabular}{l|cccc}
\hline
Domain                                                                      & Dataset   & \#Train & \#Test & Metric   \\ \hline
\multirow{2}{*}{\begin{tabular}[c]{@{}l@{}}Code \\ Generation\end{tabular}} & HumanEval &   33    &   131  & pass@1   \\
                                                                            & MBPP      &   86    &   341  & pass@1   \\ \hline
\multirow{2}{*}{\begin{tabular}[c]{@{}l@{}}Math \\ Reasoning\end{tabular}}  & MATH      &   119   &   486  & Accuracy \\
                                                                            & GSM8K     &   264   &  1055  & Accuracy \\ \hline
\begin{tabular}[c]{@{}l@{}}Reading \\ Comprehension\end{tabular}            & DROP      &   200   &   800  & F1 Score \\ \hline
Multi-task                                                                  & MMLU      &   570   &  14042 & Accuracy \\ \hline
Science                                                                     & GPQA      &   200   &  198   & Accuracy \\ \hline
\end{tabular}
\end{table}

\subsection{Baseline Configurations}
\label{asec:baseline}
We provide a detailed description of the configurations for the baseline methods, with all backbone LLMs set to \texttt{GPT-4o-mini} by default, following established practices.
\begin{enumerate}[noitemsep, topsep=0pt, leftmargin=6mm]
    \item \textbf{CoT.} It prompts an agent to decompose reasoning into sequential steps instead of producing direct answers. The implementation is from~\cite{zhang2023automatic}~\footnote{\url{https://github.com/amazon-science/auto-cot}}.
    \item \textbf{SC (CoT×5).} To improve robustness, we aggregate five CoT-generated solutions, following the approach in~\cite{zhang2025aflow,zhang_maas_2025}.
    \item \textbf{Self-Refine.} It first generates an answer using CoT reasoning and then prompts the agent to iteratively self-reflect. We allow up to five refinement iterations~\footnote{\url{https://github.com/madaan/self-refine}}.
    \item \textbf{MultiPersona.} It transforms a single LLM into multiple dynamic personas via multi-turn self-collaboration to enhance problem-solving. We use the implementation from~\cite{wang2023unleashing}~\footnote{\url{https://github.com/MikeWangWZHL/Solo-Performance-Prompting}}.
    \item \textbf{LLM-Debate.} Following~\cite{zhang_maas_2025}, we assign five LLM agents with distinct roles to engage in up to two rounds of debate, and the final decision is made through majority voting~\footnote{\url{https://github.com/composable-models/llm_multiagent_debate}}.
    \item \textbf{DyLAN.} It is a framework that dynamically selects and coordinates a team of LLM-powered agents for diverse tasks. We use the implementation from~\cite{liu2023dynamic}~\footnote{\url{https://github.com/SALT-NLP/DyLAN}}.
    \item \textbf{Plan-and-Execute.} The LangChain blog presents plan-and-execute agents that separate planning from execution to improve multi-step workflow efficiency and reduce LLM costs~\footnote{\url{https://langchain-ai.github.io/langgraph/tutorials/plan-and-execute/plan-and-execute/}}.
    \item \textbf{GPTSwarm.} It is implemented following the original settings described in~\cite{zhuge_gptswarm_2024}~\footnote{\url{https://github.com/metauto-ai/GPTSwarm}}.
    \item \textbf{ADAS.} It introduces a Meta Agent Search to iteratively program new agents based on an ever-growing archive of previous discoveries. We use the implementation from~\cite{hu_adas_2024}~\footnote{\url{https://github.com/ShengranHu/ADAS}}.
    \item \textbf{AgentSquare.} Following~\cite{zhang_maas_2025}, we adopt the modular search framework from~\cite{shangagentsquare}~\footnote{\url{https://github.com/tsinghua-fib-lab/AgentSquare}}, using early stopping with a patience of five iterations.
    \item \textbf{AFlow.} We adopt the implementation from~\cite{zhang2025aflow}~\footnote{\url{https://github.com/FoundationAgents/AFlow}}, with the maximum number of iterations set to 20 following~\cite{zhang_maas_2025}.
    \item \textbf{MaAS.} It optimizes a probabilistic supernet of agentic architectures to dynamically sample query-specific multi-agent systems. We use the official implementation from~\cite{zhang_maas_2025}~\footnote{\url{https://github.com/bingreeky/MaAS}}.
\end{enumerate}

\subsection{Implementation Details}
\label{asec:implementation}

For the flow network, the construction threshold $\delta$ is set to 0.7. The prompt generates workflows, as well as the initial system prompt and role description for each agent, as provided in Figure~\ref{afig:prompts}, and the construction process is detailed in Algorithm~\ref{alg:flownetwork}.
For the policy network used in \texttt{MANGO}, it consists of one input layer, one hidden layer with 128 neurons, and one output layer. \texttt{MANGO} is trained on the training set of each dataset. For each question, we generate 100 episodes for policy learning using the Adam optimizer with a learning rate of 0.001, a reward discount factor of 0.95, and a reward balancing coefficient $\alpha$ of 0.5. The Skip-$K$ parameter is set to 1 by default for optimal effectiveness, and its impact on efficiency is further analyzed in Appendix~\ref{asec:results}.
For textual gradient optimization, our implementation is partially from the TextGrad repository~\footnote{\url{https://github.com/zou-group/textgrad}} and adapted it to an asynchronous version, where the training episode is set to 5 with a batch size of 30. We provide the prompts used for gradient computation and updates in Figure~\ref{afig:tgprompts}.
All text embeddings, including those for flow network and RL state construction, are obtained using the \texttt{all-MiniLM-L6-v2} sentence-transformer from Hugging Face~\footnote{\url{https://huggingface.co/sentence-transformers/all-MiniLM-L6-v2}}.
We adopt \texttt{GPT-4o-mini} from the OpenAI platform as the default backbone LLM~\footnote{\url{https://platform.openai.com/docs/models/gpt-4o-mini}}. To evaluate cross-model transferability, we additionally experiment with \texttt{Qwen-2.5-72B} and \texttt{Llama-3.1-70B}, which are accessed through API calls provided by the OpenRouter~\footnote{\url{https://openrouter.ai/}}.

\SetKwInOut{KwIn}{Require}
\begin{algorithm}[t]
    \caption{The Flow Network Construction}
    \label{alg:flownetwork}
	\KwIn{
        Training tasks $\mathcal{D}$; a threshold $\delta$}
        $ctr \leftarrow 3$ \hfill $\triangleright$ \textcolor{blue}{Counter for nodes} \\ 
        \For {$i=1,2, \ldots, |\mathcal{D}|$} {
            Obtain a correct $W^{(i)} = \langle e_1^{(i)}, e_2^{(i)}, \ldots \rangle$ for task $T^{(i)} \in \mathcal{D}$

            $v' \leftarrow \emptyset$ \hfill $\triangleright$ \textcolor{blue}{Record the last node}
            
            \For{$j=1,2, \ldots, |W^{(i)}|$}{
            \uIf{$i=1$ and $j<3$}{
            $v_1.\text{addOper}(e_1^{(1)})$, $v_2.\text{addOper}(e_2^{(1)})$,
            $v_1.\text{addModel}(M_1^{(1)})$, $v_2.\text{addModel}(M_2^{(1)})$

            $\mathcal{G}.\text{addEdge}(s, v_1)$, $\mathcal{G}.\text{addEdge}(v_1,v_2)$, $v' \leftarrow v_2$
            
            \textbf{continue}
            }
            
            Recall $v_s$ and its similarity $\psi$ for $e_j^{(i)}$ on $\mathcal{G}.\mathcal{V} - v'$
            
            \uIf{$\psi < \delta$ or $v_s.M \neq M_j^{(i)}$}{
            
            $v_{ctr}.\text{addOper}(e_j^{(i)})$, $v_{ctr}.\text{addModel}(M_j^{(i)})$,
            $\mathcal{G}.\text{addEdge}(v',v_{ctr})$ if $v' \neq \emptyset$
            
            $v' \leftarrow v_{ctr}$, $ctr \leftarrow ctr + 1$
            }
            \uElse{
                $v_s.\text{addOper}(e_j^{(i)})$,
                $\mathcal{G}.\text{addEdge}(v', v_s)$ if $(v', v_s) \notin \mathcal{G}$ and $v' \neq \emptyset$,
                $v' \leftarrow v_s$
            }
            
            $\mathcal{G}.\text{addEdge}(s, v')$ if $j=1$,
            $\mathcal{G}.\text{addEdge}(v', t)$ if $j=|W^{(i)}|$ \hfill $\triangleright$ \textcolor{blue}{Build source and sink nodes}
            }
        }
Derive $P_i$, $e_i^*$, and $r_i^*$ for each node $v_i$, where $1 \le i \le |\mathcal{V}|$

\textbf{Return} Flow network $\mathcal{G}$
\end{algorithm}

\section{Additional Results}
\label{asec:results}

\begin{figure}[t]
    \centering
    \captionsetup{font=small}
    \begin{minipage}[t]{0.325\linewidth}
        \centering
        \includegraphics[width=\linewidth]{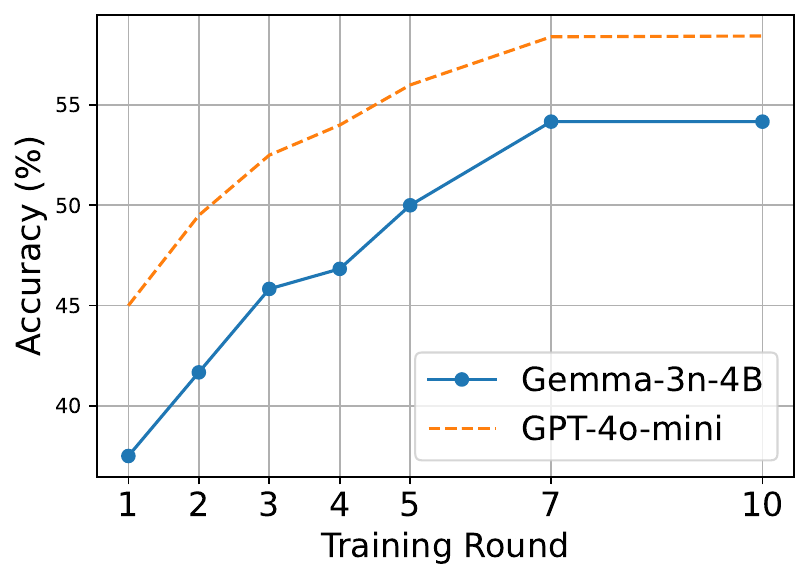}
        \caption{Flow network robustness.}
        \label{afig:4b}
    \end{minipage}
    \hfill
    \begin{minipage}[t]{0.325\linewidth}
        \centering
        \includegraphics[width=\linewidth]{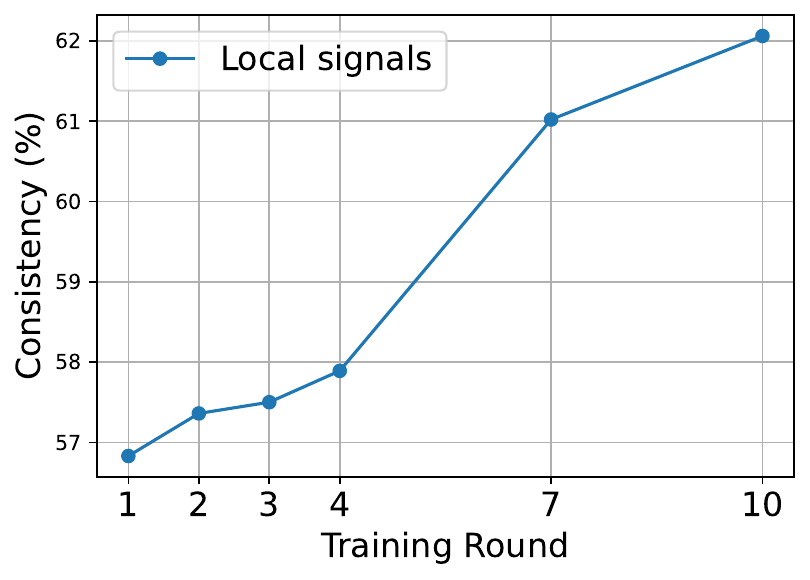}
        \caption{Local signal quality.}
        \label{afig:consistency}
    \end{minipage}
    \hfill
    \begin{minipage}[t]{0.325\linewidth}
        \centering
        \includegraphics[width=\linewidth]{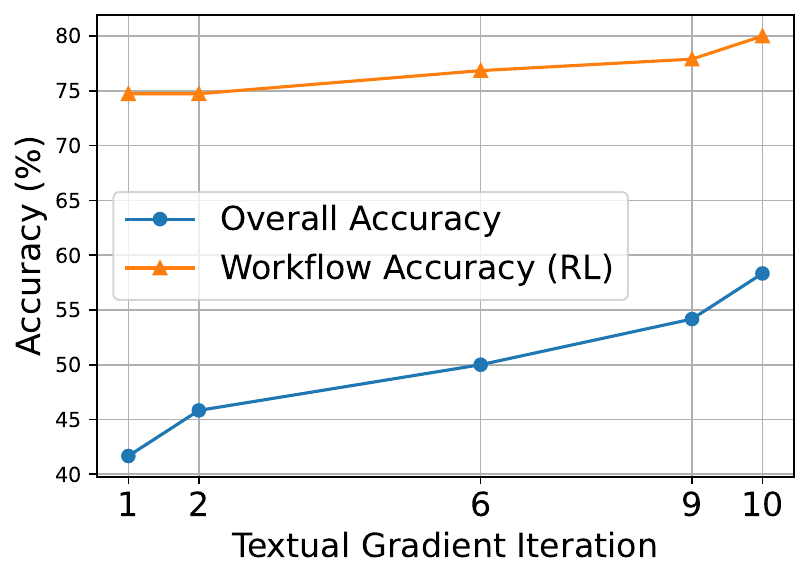}
        \caption{Joint optimization gains.}
        \label{afig:jo}
    \end{minipage}
\end{figure}

\begin{table}[t]
\centering
\caption{Robustness to erroneous workflows in the flow network.}
\label{atab:noise}
\begin{tabular}{l|cccccc}
\hline
Noise rate    & 0\% & 10\% & 20\% & 30\% & 40\% & 50\% \\ \hline
\texttt{MANGO}&58.44&56.58&54.32&53.91&53.29&52.67\\ \hline
\end{tabular}
\end{table}

\subsection{Robustness to Low-Quality Workflows in Flow Networks}
\label{asec:noise}
We study whether \texttt{MANGO} is constrained by the quality of initial workflows and how it bootstraps performance when only weak workflows are available (e.g., using a 4B model). Although poor initial workflows can affect early-stage performance, they do not impose a hard upper bound. Instead, the framework progressively improves via iterative textual-gradient refinement and reinforcement learning over the flow network, and can further incorporate newly discovered successful workflows to enhance coverage.

To evaluate this, we conduct two experiments. First, we construct the initial flow network using \texttt{Gemma-3n-4B}, where \textbf{79.8\%} of workflows are initially unsuccessful. As shown in Figure~\ref{afig:4b}, performance steadily improves across training rounds on MATH, demonstrating strong self-bootstrapping behavior. While early accuracy ranges from 37.50\% to 45.83\%, the system eventually converges to 54.17\%. For reference, using a strong LLM (\texttt{GPT-4o-mini}) achieves 58.44\%, indicating that better initial workflows mainly improve early efficiency rather than final capacity.

Second, we study the impact of erroneous workflows in the flow network on task performance by injecting noise, i.e., failed workflows from past experiences with controlled noise rates ranging from 0\% to 50\%. As shown in Table~\ref{atab:noise}, \texttt{MANGO} exhibits strong robustness to noise; even at a 50\% noise rate, performance remains relatively stable, with a {\TBD{9.9\%}} degradation. 

\subsection{Reliability and Impact of Local Signals}
\label{asec:signal}
We study the reliability and the impact of local signals in guiding intermediate reasoning steps.

To evaluate reliability, we compare intermediate outputs against ground-truth intermediate outcomes in the dataset. Intuitively, a higher alignment indicates more effective local guidance. We use \texttt{GPT-4o-mini} as a binary judge (0/1) to measure step-level consistency across training rounds on MATH. As shown in Figure~\ref{afig:consistency}, the alignment steadily improves during training, indicating that intermediate steps become increasingly consistent with ground truth. This suggests that textual gradients can provide reliable self-supervised signals even without explicit step-level supervision.

For their impact, we note that local signals are not assumed to be perfect supervision for intermediate steps. Instead, they act as approximate proxy feedback to mitigate gradient vanishing and improve credit assignment in long reasoning chains. An ablation study in Table~\ref{tab:ablation} further confirms their importance: removing local signals decreases accuracy from 58.44\% to 54.73\%, demonstrating that even approximate step-level feedback provides a useful optimization direction.

In our framework, local signals are generated by prompting \texttt{GPT-4o-mini}, where the prompts are provided in Figure~\ref{afig:signal}.

\begin{table}[t]
\centering
\caption{Impacts of $\delta$ and efficiency on the MATH benchmark.}
\label{atab:delta}
\begin{tabular}{l|cccccc}
\hline
$\delta$         & 0.0 & 0.3 & 0.5 & 0.7 & 0.9 & 1.0 \\ \hline
Accuracy (\%)    &55.35&56.17&56.79&58.44&54.94&54.73\\
Construction (s) & 8  & 14  & 40  & 62  & 90  &  8 \\
Training (min)   & 19  & 31  & 33  & 42  & 44  &  20 \\
Inference (min)    & 8  & 12  & 12  & 12  & 13  &  9 \\
\# of nodes      & 5   & 13  & 123 & 384 & 440 &  441\\ \hline
\end{tabular}
\vspace{-3mm}
\end{table}

\begin{table}[t]
\centering
\caption{Impacts of $\alpha$ in balancing rewards.}
\label{atab:alpha}
\begin{tabular}{l|ccccc}
\hline
$\alpha$       & 0.0 &0.25 & 0.5 &0.75 & 1.0 \\ \hline
\texttt{MANGO} &53.91&55.35&58.44&57.41&56.79\\ \hline
\end{tabular}
\vspace{-3mm}
\end{table}

\begin{table*}[t]
\centering
\caption{Impact of $K$ on token usage, cost, training time, and accuracy on the MATH benchmark.}
\label{atab:skip}
\begin{tabular}{l|ccccc}
\hline
Skip-$K$              & 1 & 2 & 3 & 4 & 5\\ \hline
Total prompt token          &7, 080, 219&5, 856, 443&5, 184, 917&5, 221, 400&5, 220, 612\\
Total completion token      &1, 938, 057&1, 750, 381&1, 613, 641&1, 605, 413&1, 603, 596\\
Total cost (\$)       &2.225&1.929&1.746&1.746&1.745\\
Training (min)        &42&33&31&28&27\\
Inference (min)         &12&11&10&10&9\\
Accuracy (\%)         &58.44&54.32&53.09&52.47&52.26\\ \hline
\end{tabular}
\vspace{-3mm}
\end{table*}

\subsection{Parameter Study}
\label{asec:parameter}

In our experiments, we perform minimal empirical tuning on MATH for hyperparameter selection, and the chosen settings generalize well across all other datasets, consistently yielding strong performance as reflected in Table~\ref{tab:effectiveness_baselines}.

\textbf{Threshold $\delta$ for Flow Network Structure.}
As shown in Table~\ref{atab:delta}, the threshold $\delta$ controls how operations are clustered into nodes during flow network construction. Empirically, we find that a moderate value of {\TBD{$\delta=0.7$}} yields the best performance: a small $\delta$ over-merges semantically distinct operations, reducing workflow expressiveness; while a large $\delta$ leads to overly fragmented networks with redundant nodes, increasing routing difficulty. A moderate threshold preserves meaningful workflow structures and supports effective generalization across tasks. 

\smallskip
\noindent\textbf{Threshold $\alpha$ in Balancing Rewards.}
As shown in Table~\ref{atab:alpha}, the coefficient $\alpha$ balances process-oriented and result-oriented rewards during policy optimization. Extreme values of $\alpha$ degrade performance: emphasizing only process correctness limits adaptability, while focusing solely on final outcomes weakens structural supervision. A moderate value of {\TBD{$\alpha=0.5$}} achieves the best effectiveness by jointly encouraging structurally valid workflow generation and high-quality final task outcomes.

\smallskip
\noindent\textbf{Skip-$K$.} The Skip-$K$ controls the number of nodes that can be skipped during workflow traversal, directly trading off effectiveness and efficiency in both training and inference. As shown in Table~\ref{atab:skip}, increasing $K$ reduces token usage, API cost, and training/inference time, but gradually degrades accuracy. For example, Skip-$3$ reduces the total cost by {\TBD{21.5\%}} and the runtime by {\TBD{24.1\%}}, while incurring only a modest {\TBD{10.6\%}} drop in accuracy. This parameter provides practical flexibility for balancing accuracy and computational efficiency.

\begin{table}[t]
\centering
\caption{Sensitivity analysis of planner model size and robustness under planning noise.}
\begin{tabular}{l|cccc}
\hline
\textbf{Model size} & \texttt{GPT-4o} & \texttt{GPT-4o-mini} & \texttt{LLaMA-3-8B} & \texttt{Gemma-3n-4B} \\
\hline
Accuracy (\%) & 59.05 & 58.44 & 54.53 & 53.70 \\
\hline\hline
\textbf{Noise rate} & 0\% & 20\% & 40\% & 60\% \\
\hline
Accuracy (\%) & 58.44 & 55.14 & 54.32 & 53.50 \\
\hline
\end{tabular}
\label{atab:planner_sensitivity}
\end{table}

\subsection{Sensitivity Test of Planner Model}
\label{asec:panner}
We study the sensitivity of \texttt{MANGO} to the planner model by varying model sizes (\texttt{GPT-4o}, \texttt{GPT-4o-mini}, \texttt{LLaMA-3-8B}, \texttt{Gemma-3n-4B}) and introducing controlled noise into generated plans. As shown in Table~\ref{atab:planner_sensitivity}, larger planners yield slightly better performance on MATH (up to 59.05\%), while smaller models still maintain competitive results, indicating robustness to planner capacity.

We further inject noise into the planning process by randomly perturbing the original plans with different noise rates. Although performance degrades gradually as noise increases (from 58.44\% to 53.50\%), the overall drop remains moderate, demonstrating that \texttt{MANGO} is robust to imperfect planning signals.

\begin{table*}[t]
\centering
\caption{Performance comparison across different backbone LLMs.}
\label{atab:gpt5}
\setlength{\tabcolsep}{4pt}
\begin{tabular}{lccccccc}
\toprule
\textbf{Method}            & \textbf{HumanEval} & \textbf{MBPP} & \textbf{MATH} & \textbf{GSM8K} & \textbf{DROP} & \textbf{MMLU} & \textbf{GPQA} \\
\midrule
\texttt{GPT-4o-mini} & 87.08 & 71.83 & 46.29 & 87.45 & 68.25 & 63.44 & 28.79 \\
\texttt{GPT-4o}      & 93.13 & 76.54 & 53.30 & 92.04 & 71.30 & 83.34 & 48.99 \\
\texttt{GPT-5}       & 98.47 & 80.06 & 90.33 & 94.12 & 76.23 & 93.60 & 80.81 \\
\midrule
\texttt{MANGO (GPT-4o-mini)} & 95.42 & 85.63 & 58.44 & 94.12 & 84.75 & 81.97 & 40.40 \\
\texttt{MANGO (GPT-4o)}      & 96.18 & 87.68 & 60.08 & 94.41 & 89.51 & 89.22 & 54.04 \\
\texttt{MANGO (GPT-5)}       & 98.47 & 93.26 & 94.03 & 96.68 & 90.74 & 95.10 & 84.34 \\
\bottomrule
\end{tabular}
\vspace{-3mm}
\end{table*}

\subsection{Scaling with Stronger Backbone Models}
\label{asec:gpt5}

We study how \texttt{MANGO} behaves when instantiated with increasingly capable backbone LLMs, including \texttt{GPT-4o-mini}, \texttt{GPT-4o}, and the frontier \texttt{GPT-5}. As shown in the Table~\ref{atab:gpt5}, we observe that the fundamental capability of the backbone significantly affects overall performance. Specifically, when incorporating more powerful backbones, \texttt{MANGO} continues to provide additional gains, e.g., 60.9\% improvement with \texttt{GPT-5} and 2.8\% improvement with \texttt{GPT-4o} compared to the default \texttt{MANGO (GPT-4o-mini)} on MATH. These results indicate that \texttt{MANGO} is complementary to backbone improvements and continues to yield benefits as model capabilities scale.

\subsection{Gains from Joint Optimization}
\label{asec:tg}
We study the effect of joint optimization via textual gradients and reinforcement learning. Beyond prompt refinement, textual gradients also influence workflow generation by updating node role descriptions, which in turn define the state representation used for RL-based path selection. To verify this, we report five textual gradient episodes with overall accuracy and workflow generation accuracy (binary match with ground-truth paths) under a fixed RL policy. As shown in Figure~\ref{afig:jo}, workflow accuracy consistently improves alongside overall performance, indicating that the gains are not solely from prompt tuning.

\begin{table}[t]
\centering
\caption{Webshop benchmark results using the \texttt{GPT-3.5-turbo-0301} backbone.}
\label{atab:webshop}
\begin{tabular}{lcccc}
\toprule
\textbf{Method} & DyLAN~\cite{liu2023dynamic} & CoT~\cite{wei2022chain} & MaAS~\cite{zhang_maas_2025} & \texttt{MANGO} \\
\midrule
\textbf{Webshop} & 0.683 & 0.569 & 0.697 & \textbf{0.750} \\
\bottomrule
\end{tabular}
\end{table}

\subsection{Generalization to Interactive Environments}
\label{asec:webshop}
We evaluate the \texttt{MANGO} framework in interactive environments such as WebShop~\cite{yao2022webshop}. Following the DyLAN~\cite{liu2023dynamic} setup, we compare \texttt{MANGO} with DyLAN~\cite{liu2023dynamic}, CoT~\cite{wei2022chain}, and MaAS~\cite{zhang_maas_2025} using the \texttt{GPT-3.5-turbo-0301} backbone. As shown in Table~\ref{atab:webshop}, \texttt{MANGO} consistently outperforms all baselines, demonstrating that the proposed flow network generalizes effectively beyond reasoning tasks to interactive decision-making settings.

\begin{table*}[t]
\centering
\small
\caption{Comparison against unconstrained policy routing methods.}
\label{atab:other_benchmarks}
\setlength{\tabcolsep}{2pt}
\renewcommand{\arraystretch}{1.15}
\begin{tabular}{llcccc}
\toprule

\textbf{Backbone} & \textbf{Method}
& \textbf{HumanEval (Acc / Time)}
& \textbf{GSM8K (Acc / Time)}
& \textbf{GSM-Hard}
& \textbf{MMLU-Pro} \\

\midrule

\multirow{2}{*}{\texttt{GPT-4o-mini}}
& BiRouter~\cite{yang2026augmented}
& 91.46 / 4.3
& 94.09 / 17.5
& --
& -- \\

& \texttt{MANGO}
& \textbf{95.42 / 2.7}
& \textbf{94.12 / 10.0}
& --
& -- \\

\midrule

\multirow{2}{*}{\texttt{Llama-3.1-8B}}
& Puppeteer~\cite{dang2025multi}
& --
& --
& 48.00
& 52.00 \\

& \texttt{MANGO}
& --
& --
& \textbf{48.98}
& \textbf{53.20} \\

\bottomrule
\end{tabular}
\end{table*}

\subsection{Comparison against Unconstrained Policy Routing Methods}
\label{asec:unconstrained}
We compare \texttt{MANGO} with unconstrained policy routing methods such as Puppeteer-Mono (\texttt{Llama-3.1-8B})~\cite{dang2025multi} and BiRouter (\texttt{GPT-4o-mini})~\cite{yang2026augmented}, which directly select agents without leveraging an explicit workflow structure. In contrast, \texttt{MANGO} introduces a flow network as a structured prior over routing decisions.

As shown in Table~\ref{atab:other_benchmarks}, following their experimental settings, \texttt{MANGO} consistently achieves superior performance. We summarize three key advantages of \texttt{MANGO}: (1) improved sample efficiency by reducing exploration of infeasible or low-quality paths, (2) enhanced training stability by constraining decisions to meaningful transitions, and (3) stronger generalization via recombination of workflow substructures to form novel solutions. Moreover, the flow network remains flexible and can be dynamically extended with newly discovered trajectories, ensuring that it does not limit the discovery of new patterns.

\subsection{Case Study}
\label{asec:case}
We conduct two case studies to demonstrate the effectiveness of \texttt{MANGO} in addressing workflow generation errors and single-agent execution errors, as shown in Table~\ref{atab:wferror} and Table~\ref{atab:agenterror}, respectively.
In the first case, we consider an integer programming inequality problem from MATH. MaAS incorrectly ignores the constraint that $y$ must also be an integer like $x$ (highlighted in red), resulting in an incorrect answer of 9 possible values. In contrast, \texttt{MANGO} explicitly accounts for this integer constraint (highlighted in green), correctly handling the integer factors of 54 and producing the correct result of 4.
In the second case, although MaAS generates a correct plan (checking if each sentence starts with the word ``I''), it is executed incorrectly because the agent checks whether a sentence starts with the character ``I''. \texttt{MANGO} resolves this issue by explicitly optimizing system execution.
Additionally, we observe that \texttt{MANGO} outputs fewer tokens when solving problems, as its RL policy is trained to select an optimal path, which enhances efficiency.

Furthermore, Table~\ref{atab:case_study_workflow} presents two representative cases where initial workflows fail and are corrected through training (textual gradient refinement and reinforcement learning recombination), illustrating the effectiveness of the proposed flow network in managing and improving workflows.

\section{Limitations}
\label{asec:limit}
Despite its strong empirical results, \texttt{MANGO} still has several limitations. (1) The construction of the flow network depends on the availability and quality of historical successful workflows, which may introduce bias when these workflows are suboptimal. (2) The framework involves several hyperparameters (e.g., similarity threshold $\delta$ and Skip-$K$ range), which can influence performance and require tuning. (3) The joint optimization of reinforcement learning and textual gradients introduces additional system complexity and training instability, making reproduction more challenging compared to single-agent baselines.

We note that \textbf{these challenges have been thoroughly validated in our experiments, and the results empirically confirm their existence and impact under controlled settings.} Specifically, we include robustness studies in Figure~\ref{afig:4b} and Table~\ref{atab:noise} for (1), hyperparameter studies for $\delta$ in Table~\ref{atab:delta} and Skip-$K$ in Table~\ref{atab:skip} for (2), and we release our code and data to support reproducibility for (3).

\section{Impact Statement}

\textbf{Ethical Considerations.} The proposed \texttt{MANGO} framework raises no ethical concerns. This study focuses on improving multi-agent collaboration using publicly available datasets and synthetic workflows. No experiments involve human subjects, private user data, or sensitive information. All methods are evaluated on standard benchmarks, ensuring reproducibility and minimizing ethical risks. Therefore, there are no direct ethical concerns arising from this work.

\textbf{Societal Implications.} \texttt{MANGO} introduces a data-driven paradigm for designing multi-agent systems, enabling them to learn from historical workflows in a manner analogous to how neural networks learn from training data. By automating the discovery of effective collaboration patterns and optimizing agent behaviors, this approach enhances the reliability and efficiency of intelligent systems. Potential applications include educational tools, AI assistants, and robotics, while illustrating the broader societal impact of adaptive, self-evolving multi-agent AI.

\begin{table*}[h]
\caption{Comparison of \texttt{MANGO} and MaAS in terms of workflow generation errors on MATH. Incorrect content is highlighted in red, with the corresponding correct content highlighted in green.}
\scriptsize
\label{atab:wferror}
\begin{tabular}{p{0.48\linewidth} | p{0.48\linewidth}}
\hline
\multicolumn{2}{p{\linewidth}}{\texttt{\textbf{Question:}\newline\newline
I have a deck of $54$ cards, and I deal all of the cards to $x$ players, with each player getting $y$ cards. If $x$ is at least $2$ and $y$ is at least $5$, then how many possible values of $x$ are there?\newline
}
}
\\ \hline
{\texttt{\textbf{MANGO}\newline
Agent-1: Identify the relationship between the total number of cards, players, and cards per player, expressed as the equation \(54 = x \times y\) where \(x \geq 2\) and \(y \geq 5\). 
The relationship can be expressed as \(54 = x \times y\). Rearranging gives \(y = \frac{54}{x}\).\newline\newline
Agent-2: Determine the possible \colorbox{green!15}{integer values for \(y\) that satisfy \(y \geq 5\) and are} \newline
\colorbox{green!15}{factors of 54.} 
The factors of \(54\) are \(1, 2, 3, 6, 9, 18, 27, 54\). From these, the integer values for \(y\) that satisfy \(y \geq 5\) are \(6, 9, 18, 27, 54\).\newline\newline
Agent-3: For each valid \(y\) value, calculate the corresponding \(x\) value using \(x = \frac{54}{y}\) and check that \(x \geq 2\). 
For each valid \(y\) value:\newline\newline
1. \(y = 6\): \(x = \frac{54}{6} = 9\)\newline
2. \(y = 9\): \(x = \frac{54}{9} = 6\)\newline
3. \(y = 18\): \(x = \frac{54}{18} = 3\)\newline
4. \(y = 27\): \(x = \frac{54}{27} = 2\)\newline
5. \(y = 54\): \(x = \frac{54}{54} = 1\) (not valid since \(x < 2\))\newline\newline
\colorbox{green!15}{Valid \(x\) values are \(9, 6, 3, 2\).} \newline\newline
Agent-4: Count the distinct valid values of \(x\) obtained from the previous step. 
Thus, the possible values of \(x\) are \(2, 3, 6, 9\), giving us a total of \(4\) possible values for \(x\).\newline\newline
\colorbox{green!15}{Answer: 4}
}}
&
{\texttt{\textbf{MaAS}\newline 
IO $\rightarrow$ solution\newline
Layer-1: Generate + Programmer + GenerateCoT $\rightarrow$ new\_solution1 \newline
Layer-2: SelfRefine + Programmer + EarlyStop $\rightarrow$ new\_solution2 \newline
sc\_ensemble([solution,new\_solution1,new\_solution2]) $\rightarrow$ \colorbox{red!15}{Choose new\_solution1 as final\_answer} \newline\newline
solution:\newline
...\newline\newline
new\_solution1:\newline
1. Restatement of the Problem\newline
We have a deck of 54 cards, and we want to deal all of the cards to $x$ players, with each player receiving y cards. We are given the conditions \(x \geq 2\) and \(y \geq 5\). We need to determine how many possible values of $x$ satisfy these conditions.\newline\newline
2. Explanation of Mathematical Concepts\newline
To solve this problem, we need to understand the relationship between the total number of cards, the number of players, and the number of cards each player receives. The total number of cards dealt can be expressed as: $x\times y = 54$. We can derive the possible values for $x$ by first expressing $y$ in terms of $x: y = \frac{54}{x}$\newline\newline
3. Detailed Steps to the Solution\newline
1) **Substituting the Constraints**: Since $y$ must be at least 5, we can set up the inequality: $\frac{54}{x} \geq 5$\newline
2) **Solving the Inequality**: To solve for $x$, we can multiply both sides by $x$ (noting that x must be positive): $54 \geq 5x$. Dividing both sides by 5: $\frac{54}{5} \geq x$\newline
Calculating \(\frac{54}{5}\): $\frac{54}{5} = 10.8$\newline
Since $x$ must be an integer, we have: $x \leq 10$\newline
3) **Finding the Minimum Value of \(x\)**: We also know from the problem statement that \(x \geq 2\).\newline
4) \colorbox{red!15}{**Identifying Possible Integer Values for x**}: Thus, the possible integer values for x are:\newline
$x = 2, 3, 4, 5, 6, 7, 8, 9, 10$\newline
5) **Counting the Values**: We can count these values:\newline
- The integers from 2 to 10 inclusive are 2, 3, 4, 5, 6, 7, 8, 9, 10.\newline
- This gives us a total of \colorbox{red!15}{$10 - 2 + 1 = 9$ possible values.}\newline\newline
4. Final Answer\newline
\colorbox{red!15}{The number of possible values of x is: 9}\newline\newline
5. Significance of the Result\newline
This result indicates that there are 9 different ways to distribute 54 cards among at least 2 players, with each player receiving at least 5 cards. This can be useful in card games or scenarios where equitable distribution is necessary, ensuring that all players receive a fair number of cards.\newline\newline
new\_solution2:\newline
...\newline
}}
\\ \hline
\end{tabular}
\end{table*}

\begin{table*}[h]
\caption{Comparison of \texttt{MANGO} and MaAS in terms of single-agent execution errors on HumanEval. Incorrect content is highlighted in red, with the corresponding correct content highlighted in green.}
\scriptsize
\label{atab:agenterror}
\begin{tabular}{p{\linewidth}} 
\hline
\begin{minipage}[t]{\linewidth}
\texttt{\textbf{Question:}}
\begin{lstlisting}
def is_bored(S):
    """
    You'll be given a string of words, and your task is to count the number of boredoms.
    A boredom is a sentence that starts with the word 'I'.
    Sentences are delimited by '.', '?' or '!'.
    Examples:
    >>> is_bored("Hello world")
    0
    >>> is_bored("The sky is blue. The sun is shining. I love this weather")
    1
    """
\end{lstlisting}
\end{minipage} \\
\hline
\begin{tabular}{p{0.48\linewidth}|p{0.48\linewidth}}
\begin{minipage}[t]{\linewidth}
\texttt{\textbf{MANGO}\newline
Agent-1: Split the input string S into sentences using the delimiters '.', '?', and '!'.}
\begin{lstlisting}
sentences = [s.strip() for s in re.split(r'[.?!]', S) if s]
\end{lstlisting}

\texttt{Agent-2: Count the number of \colorbox{green!15}{sentences that start with} \newline 
\colorbox{green!15}{the word "I".}}
\begin{lstlisting}
boredom_count = sum(s.startswith("I ") or s == "I" for s in sentences)
\end{lstlisting}

\texttt{Answer:}
\begin{lstlisting}
def is_bored(S):
    sentences = [s.strip() for s in re.split(r'[.?!]', S) if s]
    boredom_count = sum(s.startswith("I ") or s == "I" for s in sentences)
    return boredom_count
\end{lstlisting}
\texttt{Evaluation:}
\begin{lstlisting}
~\good{>>> is\_bored("Is the sky blue?")}~
~\good{0}~
\end{lstlisting}
\end{minipage}
&
\begin{minipage}[t]{\linewidth}
\texttt{\textbf{MaAS}\newline 
Layer-1: Generate $\rightarrow$ current solution:\newline 
1. Split the input string into sentences using the delimiters '.', '?', and '!'.\newline 
2. Trim any leading or trailing whitespace from each sentence.\newline 
3. Check if each sentence starts with the word "I".\newline
4. Count how many sentences meet this criterion and return that count.}
\begin{lstlisting}
def is_bored(S):
    # Split the string into sentences
    import re
    sentences = re.split(r'[.!?]', S)
    # Initialize a counter for boredoms
    boredom_count = 0
    # Check each sentence
    for sentence in sentences:
        # Strip leading and trailing whitespace
        trimmed_sentence = sentence.strip()
        ~\bad{\# Check if the sentence starts with "I"}~
        if trimmed_sentence.startswith("I"):
            boredom_count += 1
    return boredom_count
\end{lstlisting}
\texttt{Layer-2: Generate + test $\rightarrow$ current solution:\newline 
To implement the 'is\_bored' function, we need to follow these steps:\newline 
1. Split the input string into sentences \newline 
2. Trim any leading or trailing whitespace from each sentence.\newline 
3. Count how many of these sentences start with the word "I".\newline\newline 
test:}
\begin{lstlisting}
>>> is_bored("Hello world") 
0
>>> is_bored("The sky is blue. The sun is shining. I love this weather.")
1
\end{lstlisting}
\begin{lstlisting}
def is_bored(S):
    # Split the string into sentences
    import re
    sentences = re.split(r'[.!?]', S)
    # Initialize a counter for boredoms
    boredom_count = 0
    # Check each sentence
    for sentence in sentences:
        # Strip leading and trailing whitespace
        trimmed_sentence = sentence.strip()
        # Check if the sentence starts with "I"
        if trimmed_sentence.startswith("I"):
            boredom_count += 1
    return boredom_count
\end{lstlisting}

\texttt{Layer-3: Early-Exit\newline\newline
Evaluation:}
\begin{lstlisting}
~\bad{>>> is\_bored("Is the sky blue?")}~
~\bad{1}~
\end{lstlisting}
\end{minipage}
\end{tabular} \\
\hline
\end{tabular}
\end{table*}

\begin{table*}[h]
\caption{Case studies showing how \texttt{MANGO} corrects erroneous workflows. Incorrect and correct contents are highlighted in red and green, respectively.}
\setlength{\tabcolsep}{1pt}
\begin{tabular}{
>{\raggedright\arraybackslash}p{2cm}
|>{\raggedright\arraybackslash}p{4cm}
|>{\raggedright\arraybackslash}p{4cm}
|>{\raggedright\arraybackslash}p{4.5cm}
}
\toprule
\textbf{Question} & \textbf{Wrong workflow in the initial network} & \textbf{Correct workflow for the question after training} & \textbf{Explanation} \\
\midrule

\textbf{Q1}: Estimate $14.7923412^2+13.8914535^2$ to the nearest integer? &
\begin{minipage}[t]{\linewidth}
\raggedright
\colorbox{red!15}{Step-1}: Round the number 14.7923412 and 13.8914535 to the nearest integers. \\
\colorbox{gray!30}{Step-2}: Square two rounded numbers. \\
\colorbox{gray!30}{Step-3}: Add the two numbers together to get the answer.\newline\newline
\colorbox{red!15}{Wrong answer: 421} (Wrong for estimating the number to the nearest integer in \colorbox{red!15}{Step-1})
\end{minipage}
&
\begin{minipage}[t]{\linewidth}
\raggedright
\colorbox{green!15}{Step-1}: Round 14.7923412 to a nearby easy-to-square number (e.g., 14.79). \\
\colorbox{gray!30}{Step-2}: Square two rounded numbers. \\ 
\colorbox{gray!30}{Step-3}: Add the two numbers together to get the answer.\newline\newline
\colorbox{green!15}{Correct answer: 412}
\end{minipage}
&
In our flow network, we observe that the incorrect \colorbox{red!15}{Step-1} is clustered with a similar operation node labeled ``Round a decimal number to a nearby easy-to-square number''. This triggers the \textbf{textual gradient} to revise the system prompts within the node, making it aware that \textbf{rounding should not be to an integer for squaring}. As a result, it produces the correct \colorbox{green!15}{Step-1} even the initial is wrong in solving Q1.
\\
\midrule

\textbf{Q2}: How many whole numbers between 99 and 999 contain exactly one 0? &
\begin{minipage}[t]{\linewidth}
\raggedright
\colorbox{gray!30}{Step-1}: Identify the range of three-digit whole numbers between 99 and 999. \\ 
\colorbox{gray!30}{Step-2}: Calculate the valid combinations for each scenario where ``0'' is placed, ensuring the other digits are not ``0''. \\ 
\colorbox{red!15}{Step-3}: Sum the valid combinations to find the total count of whole numbers that meet the criteria.\newline\newline
\colorbox{red!15}{Wrong answer: 243} (Wrong for not removing the cases where the hundreds digit is 0 in \colorbox{red!15}{Step-3})
\end{minipage}
&
\begin{minipage}[t]{\linewidth}
\raggedright
\colorbox{gray!30}{Step-1}: Identify the range of three-digit whole numbers between 99 and 999. \\ 
\colorbox{gray!30}{Step-2}: Calculate the valid combinations for each scenario where ``0'' is placed, ensuring the other digits are not ``0''. \\ 
\colorbox{green!15}{Step-3}: Ensure the highest bit is not 0 and calculate the total count.\newline\newline
\colorbox{green!15}{Correct answer: 162}
\end{minipage}
&
By examining flow network structure, we observe that Step-2 of the incorrect workflow is merged with another workflow, creating a branching node. One branch leads to a node containing the incorrect \colorbox{red!15}{Step-3}, while the other leads to a node labeled ``Ensure the highest digit is not 0 when forming a number'' (i.e., \colorbox{green!15}{Step-3}). Through the \textbf{RL-based training}, the solution for Q2 gradually learns to follow the other branch, where the \textbf{constraint on the highest digit being non-zero is considered}.
\\
\bottomrule
\end{tabular}
\label{atab:case_study_workflow}
\end{table*}

\clearpage

\begin{figure*}[t]
\centering
{\arrayrulecolor{black}
\begin{minipage}{\linewidth}
\setlength{\arrayrulewidth}{2pt}  
\begin{tabular}{|p{\textwidth}|}
\hline
\rowcolor{brown}
\parbox{11cm}{\vspace{1mm} {\color{white} \large Workflow Generation, Initial System Prompt and Role Description}}\vspace{1mm}\\
\texttt{You are an assistant. Provide workflow steps for tasks without test and verify process. For each step, the content and role description must follow the following format:\newline\newline
\{\newline 
\hspace*{1em} (<serial number>)Content: <abstract content> | <specific content>.\newline 
\hspace*{1em} Role Description: <role> | <related description>.(/<serial number>)\newline 
\}\newline\newline
The " | " part must be included. The "abstract content" refers to the abstract action of the step. The "specific content" refers to specific content of the step.\newline\newline 
For the task: \{task\}\newline 
Based on the correct answer: \{gt\}\newline
Generate a clear and concise workflow consisting of several steps. Produce a "minimal-sufficient" workflow that completes the task with the fewest steps. Each step should be actionable and ordered sequentially. For each step, give a role description that can perform this kind of things.\newline\newline\newline
SYSTEM\_PROMPT = f"Target: You are a \{role\_description\}. Execute the CURRENT\_STEP without extra text, explanation, comment."\newline
Operating Principles: Focus only on the CURRENT\_STEP; do not expand scope, jump ahead or solve the total task by yourself. Double-check your calculations or reasoning.
} \\ \hline
\end{tabular}
\end{minipage}
}
\caption{Prompts used in flow network construction.}
\label{afig:prompts}
\end{figure*}

\begin{figure*}[t]
\centering
{\arrayrulecolor{black}
\begin{minipage}{\linewidth}
\setlength{\arrayrulewidth}{2pt}  
\begin{tabular}{|p{\textwidth}|}
\hline
\rowcolor{brown}
\parbox{11cm}{\vspace{1mm} {\color{white} \large Planner Model}}\vspace{1mm}\\
\texttt{GIVEN\_TASK: \{given\_task\}\newline\newline  
For the GIVEN\_TASK, generate a clear and concise workflow consisting of 1 to \{max\_step\} steps. Add an opening tag "(<serial num>)" and a closing tag "(/<serial num>)" for each step like this:\newline\newline
(1) STEP\_TEXT (/1)\newline\newline
(2) STEP\_TEXT (/2)\newline\newline
...\newline\newline
Do not provide final answer.\newline
} \\ \hline
\end{tabular}
\end{minipage}
}
\caption{Prompts used in planner model.}
\label{afig:planprompts}
\end{figure*}

\begin{figure*}[t]
\centering
{\arrayrulecolor{black}
\begin{minipage}{\linewidth}
\setlength{\arrayrulewidth}{2pt}  
\begin{tabular}{|p{\textwidth}|}
\hline
\rowcolor{brown}
\parbox{8cm}{\vspace{1mm} {\color{white} \large Local Signal Generation}}\vspace{1mm}\\
\texttt{You are part of an optimization system that improves a given text (i.e. the variable). You are the gradient (feedback) engine.\newline\newline
Your only responsibility is to give intelligent and creative feedback and constructive criticism to variables, given an objective specified in <OBJECTIVE\_FUNCTION> </OBJECTIVE\_FUNCTION> tags.\newline\newline
The variables may be solutions to problems, prompts to language models, code, or any other text-based variable.\newline\newline
Pay attention to the role description of the variable, and the context in which it is used. You should assume that the variable will be used in a similar context in the future.\newline\newline
Only provide strategies, explanations, and methods to change in the variable. DO NOT propose a new version of the variable, that will be the job of the optimizer. Your only job is to send feedback and criticism (compute "gradients").\newline\newline
For instance, feedback can be in the form of "Since language models have the X failure mode...", "Adding X can fix this error because...", "Removing X can improve the objective function because...", "Changing X to Y would fix the mistake ...", that gets at the downstream objective.\newline\newline
If a variable is already working well (e.g. the objective function is perfect, an evaluation shows the response is accurate), you should not give feedback.
} \\ \hline
\end{tabular}
\end{minipage}}
\caption{Prompts used in generating local signals.}
\label{afig:signal}
\end{figure*}

\begin{figure*}[t]
\centering
{\arrayrulecolor{black}
\begin{minipage}{\linewidth}
\setlength{\arrayrulewidth}{2pt}  
\begin{tabular}{|p{\textwidth}|}
\hline
\rowcolor{brown}
\parbox{8cm}{\vspace{1mm} {\color{white} \large Gradient Computation and Updates}}\vspace{1mm}\\
\texttt{\# Execute (run/call) a loss function and compute the loss value\newline
You are a smart language model that evaluates the response for the task. You do not solve task or propose new responses, only evaluate model response critically and give very concise feedback.\newline\newline
GLOSSARY\_TEXT = """\newline
\#\#\# Glossary of tags that will be sent to you:\newline
\# - <LM\_SYSTEM\_PROMPT>: The system prompt for the language model.\newline
\# - <LM\_INPUT>: The input to the language model.\newline
\# - <LM\_OUTPUT>: The output of the language model.\newline
\# - <FEEDBACK>: The feedback to the variable.\newline
\# - <CONVERSATION>: The conversation history.\newline
\# - <FOCUS>: The focus of the optimization.\newline
\# - <ROLE>: The role description of the variable.\newline"""\newline\newline
\# System prompt to TGD\newline
OPTIMIZER\_SYSTEM\_PROMPT = (\newline
    \hspace*{1em}"You are part of an optimization system that improves text (i.e., variable). "\newline
    \hspace*{1em}"You will be asked to creatively and critically improve prompts, solutions to problems, code, or any other text-based variable."\newline
    \hspace*{1em}"You will receive some feedback, and use the feedback to improve the variable. "\newline
    \hspace*{1em}"The feedback may be noisy, identify what is important and what is correct. "\newline
    \hspace*{1em}"Pay attention to the role description of the variable, and the context in which it is used. "\newline
    \hspace*{1em}"This is very important: You MUST give your response by sending the improved variable between \{new\_variable\_start\_tag\}\{\{improved variable\}\} \{new\_variable\_end\_tag\} tags."\newline
    \hspace*{1em}"The text you send between the tags will directly replace the variable."\newline
    \hspace*{1em}f"\{GLOSSARY\_TEXT\}"\newline
)} \\ \hline
\end{tabular}
\end{minipage}}
\caption{Prompts used in textual gradient descent.}
\label{afig:tgprompts}
\end{figure*}

\end{document}